\documentclass[authoryear,preprint,review,12pt]{elsarticle}

\usepackage{amssymb}
\usepackage{amsmath}
\usepackage{hyperref}
\usepackage{multicol}
\usepackage{array, booktabs, caption}
\usepackage{subcaption}

\usepackage{listings}

\newcolumntype{C}[1]{>{\centering\let\newline\\\arraybackslash\hspace{0pt}}m{#1}}

\journal{Journal of Memory and Language}

\begin{document}

\begin{frontmatter}

\title{LLMs model how humans induce logically structured rules} %

\author[inst1,inst2]{Alyssa Loo}
\ead{alyssa_loo@alumni.brown.edu}
\author[inst1,inst2]{Ellie Pavlick}
\ead{ellie_pavlick@brown.edu}
\author[inst3,inst2]{Roman Feiman\corref{cor1}}
\ead{roman_feiman@brown.edu}

\affiliation[inst1]{country={Department of Computer Science, Brown University, Providence, RI, USA}
            }
\affiliation[inst2]{country={Program in Linguistics, Brown University, Providence, RI, USA}
            }
\affiliation[inst3]{country={Department of Cognitive and Psychological Sciences, Brown University, Providence, RI, USA}
            }
            
\cortext[cor1]{Corresponding Author. Correspondence should be directed to Department of Cognitive and Psychological Sciences, Brown University
Room 241, 190 Thayer St, Providence, RI, or the corresponding author's email.}

\address{Brown University, Providence, Rhode Island, USA}
\begin{abstract}
A central goal of cognitive science is to provide a computationally explicit account of both the structure of the mind and its development: what are the primitive representational building blocks of cognition, what are the rules via which those primitives combine, and where do these primitives and rules come from in the first place? A long-standing debate concerns the adequacy of artificial neural networks as computational models that can answer these questions, in particular in domains related to abstract cognitive function, such as language and logic. This paper argues that recent advances in neural networks -- specifically, the advent of large language models (LLMs) -- represent an important shift in this debate. We test a variety of LLMs on an existing experimental paradigm used for studying the induction of rules formulated over logical concepts. Across four experiments, we find converging empirical evidence that LLMs provide at least as good a fit to human behavior as models that implement a Bayesian probablistic language of thought (pLoT), which have been the best computational models of human behavior on the same task. Moreover, we show that the LLMs make qualitatively different predictions about the nature of the rules that are inferred and deployed in order to complete the task, indicating that the LLM is unlikely to be a mere implementation of the pLoT solution. Based on these results, we argue that LLMs may instantiate a novel theoretical account of the primitive representations and computations necessary to explain human logical concepts, with which future work in cognitive science should engage.
\end{abstract}


\end{frontmatter}

\section{Introduction}
\label{sec:introduction}

Can artificial neural networks (ANNs) provide an adequate model of higher cognition? The debate over this question goes back to the connectionist models of the 1980s \citep{rumelhartParallelDistributedProcessing1987, fodorConnectionismCognitiveArchitecture1988a}, but has been renewed by advances in the behavioral capabilities of these systems \citep{mcgrath2024DNNsInformTheory}. On the one hand, ANNs, which interact with the world by mapping a set of input features to a range of output behaviors, have several attractive properties as candidate cognitive models. Their behaviors are learned through simple associative feedback mechanisms, meaning that any capabilities they can be shown to achieve, as well as any representations that support those capabilities, are inherently paired with a mechanistic, demonstrably implementable account of acquisition. They typically operate over inputs that are ``subsymbolic’’, such as raw pixels in the case of vision or raw soundwaves or text characters in the case of speech and language, meaning that they do not require a theoretical commitment to any complex inventory of custom-built, content-specific innate concepts \citep{fodorLanguageThought1975,fodor2008lot2}. And, being at least loosely inspired by biological neural networks \citep{rumelhartParallelDistributedProcessing1987}, ANNs seem to bring us closer to an implementation-level account of cognition than alternative computational models, such as those built on combinatorial symbolic structures. On the other hand, ANN models of cognition rarely live up to their billing in practice: the primitive symbols that ANNs take as input are often not theory-neutral, and require hand-crafted encodings; the account of acquisition that ANNs model is not necessarily a plausible one for humans \citep{warstadtFindingsBabyLMChallenge2023,javier-vazquez-martinez-etal-2023-evaluating}; and while ANNs were originally developed by cognitive scientists, modern architectures and learning algorithms draw only scant inspiration from the human brain \citep{Goodfellow-et-al-2016}. 

But even with these shortcomings, a good ANN model of human cognitive function would be a theoretical achievement. It would serve as an existence proof that there is one common computational architecture that provides a unified answer to three fundamental and deeply intertwined questions in cognitive science: What are the representational primitives that underlie cognition? What are the computational principles via which these primitives are combined? And, how are these computational principles acquired?

Nevertheless, a widely-held view is that ANNs can never be adequate models of higher cognition. Arguments are typically made on two grounds. First, that ANNs in principle do not have the right type of architecture to explain cognition. For instance, that ANNs lack innate representational support for key properties of symbolic computation, such as content-independence, compositionality, and variable binding, which humans are argued to have \citep{newell1980physical,marcusAlgebraicMindIntegrating2001, quilty-dunnBestGameTown2023a}. Second, that ANNs are, in practice, empirically inadequate as models of human cognition. This is supported by the many studies in which ANNs demonstrate inefficient learning and poor generalization \citep{kaplan2020scaling,linzen2020can,zhang2021you} compared to humans. These two arguments are highly interconnected, as ANNs’ empirical inadequacy is taken as evidence of the inherent limitations of their architecture: Whatever the representational primitives and computational principles underlying human cognition are, ANNs do not have them innately; and the  empirical inadequacy of fully trained ANNs casts doubt on whether ANNs' learning procedure can acquire them from data.

Logic is a key example of a domain in which generic ANNs have been argued to be empirically inadequate as models of cognition, presumably because they fundamentally lack the right structure \citep{fodorConnectionismCognitiveArchitecture1988a,quilty-dunnBestGameTown2023a}. In the paradigm case, logical reasoning uses a set of primitive operators -- e.g., boolean AND, OR, and NOT -- as well as abstract combinatorial mechanisms such as variable binding, which can be deployed to compose the same operators with different content. Although which exact logical operators humans reason with has long been an open question, modeling them with even just simple Boolean primitives has proven to be a good fit for human behavior on certain reasoning tasks within controlled settings \citep{bruner1956study,feldman2000minimization,feldmanSimplicityPrincipleHuman2003,kemp2012exploring,piantadosiLogicalPrimitivesThought2016}. Neural networks that lack such structure have struggled to demonstrate that they warrant serious consideration as models of cognition in these domains \citep{ellisHumanlikeFewShotLearning2023,sable2021sensitivity,dehaene2022symbols}.

There is no question about whether neural networks are theoretically capable of implementing the right sort of structures for logical reasoning. This has long been conceded in principle \citep{fodorConnectionismCognitiveArchitecture1988a}, since ANNs were shown to be able to implement Boolean logic gates early on \citep{mcculloch1943logical}, and were subsequently proven to be Turing complete \citep{pollack1987connectionist,siegelmann1992computational}. It has more recently been demonstrated empirically. For example, work that builds compositionality and variable binding into the training curriculum \citep{lake2023human} or directly into the neural architecture \citep{smolensky2022neurocompositional} has demonstrated greatly improved capacity for compositional generalization. Other work has demonstrated that, when exposed to specialized curricula or training signals, models can learn to behave like systems that have significant built-in internal structure, for example, simulating Bayesian networks \citep{mccoy2023modeling} or logical theorem provers \citep{evans2018can}. Parallel lines of work have shown that rules governed by more complex logical operators take proportionally more evidence to learn, both for symbolic systems with pre-specified logical primitives \citep{carcassi2023boolean} and for neural networks \citep{steinert2019learnability,carcassi2021monotone,carcassi2022neural}. Such results make a theoretical contribution that obeys a division of labor between levels of analysis \citep{marr1982vision}: The symbolic model provides the core computational-level theory (What are the primitives? What are the rules?) and the ANN model provides an account of how this architecture might be implemented.

What remains elusive, however, is the hypothetical ANN that achieves human-like competence in logical reasoning \textit{without} its designers building in logical primitives or computational principles \textit{a priori} through either the architecture or the training procedure. From a theoretical perspective, such an ANN differs in two important ways from the ``mere implementations'' \citep{fodorConnectionismCognitiveArchitecture1988a} of symbolic accounts, described above. First, by not building the logical primitives or computational rules into the design of the ANN, the ANN has the potential to model not only the primitives and the compositional rules, but also a way that both primitives and rules can be acquired. Second, such an ANN allows for the possibility of primitives and rules which differ in theoretically interesting ways from classical logical primitives. That is, an unconstrained ANN could learn to solve logical reasoning problems by employing operators or rules that differ from those formalized in Boolean or first-order logic in a number of ways. These `ANN-discovered' logic-like operators may perform some, but not all of the functional role of the classical logical operators; or they may work the same way as classical operators, but over a more limited domain of inputs; or they may not be entirely ``formal'', such that the conclusions that they license depend on the content with which they compose, in addition to the logical form of the inference; In the extreme case, ANNs may discover and deploy reasoning mechanisms that are not even logic-like, but work on entirely different principles, such as resemblance or co-occurrence.

These kinds of differences may prove essential for a model of human cognition. Human reasoning behavior has frequently been shown to diverge from formal logic, exhibiting pervasive and systematic errors. This has led to a longstanding theoretical divide between two views of how humans naturally reason. 
On one side, many researchers have argued that human reasoning does not naturally use formal logical operators, at least in the absence of slow and deliberate thinking, special talent, or training \citep{kahneman1972subjective,cheng1985pragmatic,cosmides1989logic,johnson1989mental,evans2002logic,deneys2023further}. For example, people reason differently when formally equivalent cases are presented in different ways \citep{cheng1985pragmatic,cosmides1989logic,evans2002logic}, and tend to endorse invalid reasoning if it leads to conclusions they already believe \citep{evans1983conflict,klauer2000belief,cohen2017beliefs}. This type of evidence has led many researchers to suggest that ordinary, intuitive human reasoning, unlike logic, does not at its core separate the form from the content of inferences, and may more generally operate on associative \citep{sloman1996empirical,kahneman2011thinking} or social \citep{mercier2011humans} principles that are not primarily concerned with producing true beliefs. If ANNs that have not been designed to implement a formal logic can learn to match human reasoning behavior, and if they do that by discovering reasoning primitives and procedures that are inherently not formal or logical, that would bolster the view that human reasoning uses such primitives and procedures as well. 

On the other side of the divide, researchers have defended the view that human reasoning does use formal logical operators, despite the biases and errors. They argue that reasoning errors may not be due to a lack of logical competence, but to other factors, such as resource constraints on what information is found or selected \citep{newell1972human,oaksford2007bayesian,ragni2018selecting,lieder2020resource}, or the pragmatic mechanisms that mediate between logical thought and the interpretation of language -- including the language used to communicate reasoning tasks \citep{grice1975logic,braine1990natural,tessler2022logic}. On this view, however, there are still open questions about what the primitive logical concepts actually are and how they are expressed in language. For any given logical element in language and thought, there are multiple alternative formal accounts; see, for example, book-length treatments of the meaning of ``if'' \citep{khoo2022meaning}, or of negation \citep{horn1989natural}. There are likewise open questions about how (and whether) logical concepts are acquired \citep[see, e.g.][on the acquisition of negation]{mcdermott2025development}. On this side of the divide over the nature of reasoning, ANNs may be informative as well. If ANNs that have not been designed to implement a formal logic can learn to match human reasoning behavior, and if they do that by discovering formal logical operators and rules, then characterizing these could not only help to choose between existing hypotheses about the formal operators humans reason with, but could also produce new hypotheses.

Thus, an ANN model that can explain human behavior as well or better than existing models, without presupposing a specific set of logical primitives or compositional rules, has the potential to be a fundamentally new model of human logical reasoning. Such a model could be more than an implementation-level theory of a known cognitive model. It could be an implementation of a new cognitive theory that, in addition to being accompanied by an account of acquisition, could also differ in meaningful ways from existing theories in how it defines at least one of the primitives or the rules.

Large language models (LLMs) \cite{radfordLanguageModelsAre2019} represent a recent and significant advance in the empirical performance of ANNs, which might embody exactly this type of new cognitive theory. LLMs, in many respects, preserve the basic principles of ANNs that have been central to debates about their ability to model cognition since the 1980s. They possess a relatively minimalist neural network architecture, which does not contain any explicit inductive bias for logical reasoning. They are based on a powerful but generic architecture, the Transformer \citep{vaswani2017attention}, and are trained on a generic language modeling objective --  predicting the next word in a sentence -- which does not inherently encode any specialized signal for logical reasoning. %
What makes LLMs different from earlier models is primarily scale: LLMs are generic architectures trained on generic predictive modeling tasks, but are significantly larger, by many orders of magnitude, than the models that had been studied before. Thus, if LLMs exhibit a capacity for logical reasoning, it warrants a genuine reassessment of many of the theoretical and empirical arguments that have been made previously about the role that ANNs should play in computational cognitive theory.

\subsection{Logical Reasoning in LLMs}
LLMs' capacity for logical reasoning is relevant for cognitive theory if two conditions hold. First, if there is evidence that LLMs can model human behavioral data as well or better than existing (symbolic) accounts of logical reasoning from computational cognitive science. Second, if there is evidence that the LLM does not serve to merely implement these symbolic accounts, but rather differs from them in theoretically interesting ways. If both of these conditions hold, then LLMs instantiate a competing theory of human logical reasoning that must be taken seriously and investigated further by cognitive scientists. 

There has been extensive work evaluating the ability of LLMs to reason against a formal logical standard \citep{hanFOLIONaturalLanguage2024, srivastavaImitationGameQuantifying2023, wu-etal-2023-hence, betzCriticalThinkingLanguage2020, saparovLanguageModelsAre2023, saparovTestingGeneralDeductive2023, baoAssessingEnhancingRobustness2024, huaDisentanglingLogicRole2024}. Much of this work has documented specific contexts and task formulations under which LLMs succeed (e.g. \citet{weiChainofThoughtPromptingElicits2023} or fail (e.g. \citet{ettingerWhatBERTNot2020, chadihelweReasoningTransformerbasedModels2021a} at tasks that appear to require logical reasoning, without specifying the underlying representations or processes that support the successes or explain the failures.

A smaller number of studies have directly compared the performance of LLMs and humans on logical reasoning tasks \citep{andoEvaluatingLargeLanguage2023a,lampinen2024language,eisapeSystematicComparisonSyllogistic2024}, finding suggestive convergences in patterns of successes and failures, including similar ``content effects''. However, if the reason people make these types of errors is interference from other cognitive processes, such as biases and heuristics, then these errors may not indicate a lack of logical competence \citep{feiman2023conflict}. It is then likewise unclear whether LLMs' reasoning errors reflect similar interference, or a more fundamental entanglement between content and logic-like operators in their core reasoning competence.

We test whether LLMs have acquired human-like logical concepts more directly. We do not test whether they are capable of reasoning correctly from logically structured thoughts (e.g., by completing syllogisms), but rather whether they are capable of forming such thoughts in the first place.

\subsection{Rule Learning Paradigm}
We use a \textit{logical rule learning} paradigm that has been used extensively with humans \citep{bruner1956study,feldman2000minimization,feldmanSimplicityPrincipleHuman2003,kemp2012exploring,piantadosiLogicalPrimitivesThought2016}, and which offers access to human learning curves as well as error patterns, against which computational models can be compared. This paradigm presents participants with sets of objects, asks them to induce an underlying rule, and then use this rule to categorize new exemplars (e.g. what makes an object ``wudsy"?)  (Figure \ref{fig:samplegraph}). The objects are made 
 up of combinations of just a few different primitive features (e.g. three colors, three shapes, three sizes). The rule governing what counts as ``wudsy'' is formulated in terms of different logical combinations of these features, ranging from the simple to the complex in different runs of the task. These rules are not known to participants, and thus must be inferred from examples. The learning trajectory that participants follow, the mistakes they make, and the relative ease of learning different rules have all been used as signals of the process of human inductive learning and informative clues to the logical concepts they deploy  \citep{feldmanSimplicityPrincipleHuman2003}.  
 
When computational models of this rule learning paradigm have been evaluated on whether they faithfully reproduce these learning signatures, Bayesian models implementing a probabilistic Language of Thought (pLoT) have far outstripped any alternative accounts from similarity-based, rule-based or prototype-based learning \citep{piantadosiLogicalPrimitivesThought2016, goodmanRationalAnalysisRuleBased2008,sable2021sensitivity,dehaene2022symbols}. These Bayesian models are initialized with libraries of logical primitives, and use probabilistic methods to iteratively generate and score hypotheses to account for observed data. \citet{piantadosiLogicalPrimitivesThought2016} initialized such Bayesian models with different primitives and compared them for fit to human learning, enabling the authors to test which sets of logical concepts and which ways of combining them most closely match human behavior on this task. 

In contrast to these Bayesian pLoT models, the LLMs we test here have no innately specified logical concepts, nor are they explicitly programmed with a probabilistic inference procedure. %
We start by testing whether LLMs can succeed on this logical rule learning task, both in absolute terms and relative to humans. If LLMs achieve human-level success on the logical rule-learning task, we aim to investigate if how they perform the task is human-like. If LLMs then demonstrate reasoning behaviors and learning signatures that are similar to humans, that would constitute behavioral evidence that LLMs are using similar logical (or logic-like) operators as those employed by humans performing this task.

\section{Experiment 1: Models' Task Performance}
We first investigate which LLMs, if any, succeed on the logical rule learning task in their off-the-shelf pretrained state. We test a range of recent models across different sizes and in both chat and completion formats. Since each LLM is an artifact of its stochastic hill-climbing procedure from its pretraining, we might not expect every model to succeed. Our aim in Experiment 1 is to test whether any model does. If we identify any successful models, we can then further test \textit{how} they succeed—whether they are using logical (or logic-like) concepts similarly to those people use on this task.

As an additional benefit, the logical rule learning task makes use of stimuli that are unlikely to resemble an LLM's naturally occurring training data \citep{mitchellAbstractionAnalogyMakingArtificial2021}, making it less susceptible to concerns related to training task artifacts and interference \citep{mccoyEmbersAutoregressionUnderstanding2023}. While  \citet{ellisHumanlikeFewShotLearning2023} had employed LLMs on this same task, our work uses the LLM as the end-to-end inference process without offloading any component to an external symbolic reasoner, and explores tuning the LLM's weights instead of only using pretrained models.
\subsection{Method} 
\paragraph{Data Availability} Data and code for all experiments are available in \href{https://github.com/mariebiscuit/calibration-rule-learning}{this GitHub repository}.

\begin{figure}[h!]
\centering
\includegraphics[width=\textwidth]{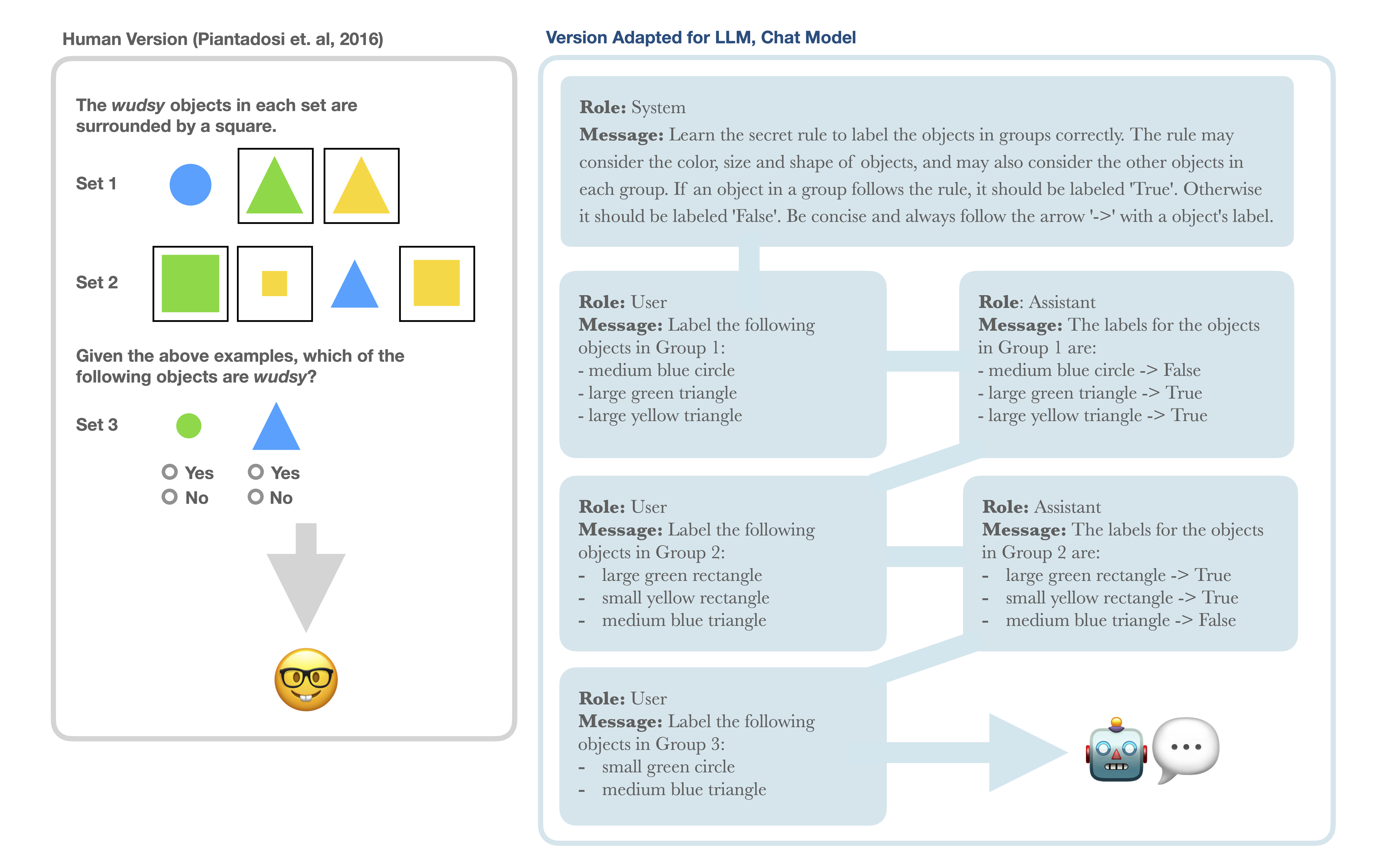}
\caption{Logical rule learning task. Sample inputs compare the original presentation to human participants in \citet{piantadosiLogicalPrimitivesThought2016} (`PTG16') versus our adapted format for a chat completion model. Inputs are constructed by concatenating instruction text with an object list and its corresponding labels according to the rule. In this example, the rule is `\texttt{same shape as a yellow object}'}
\label{fig:samplegraph}
\end{figure}

\label{sec:exp1method}
\paragraph{Task} We adapted the rule-learning experiment of \citet{piantadosiLogicalPrimitivesThought2016}, henceforth \textit{`PTG16'}. In that experiment, each participant was instructed to learn the meaning of a nonce adjective that described some visual objects, but not others. One to five objects were shown at a time as a set, and participants were told to select only the objects described by the nonce adjective. As objects varied along the dimensions of size (\textit{small, medium, large}), color (\textit{blue, green, red}) and shape (\textit{circle, triangle, square}), the meaning of the nonce adjective could be a concept pertaining to a feature dimension (e.g. ``An object is `wudsy' if it is blue"), a combination of feature dimensions (e.g. ``An object is `wudsy' if it is blue or small, but not both."), or how objects' features related to other objects in the set (e.g. ``An object is `wudsy' if it is the only medium-sized object in its set"). Rules formulated over combinations of features could always be formulated in terms of combinations of Boolean operators (\textit{and, or, not}) in a Propositional Logic. Rules on which whether one object is `wudsy' depended on properties of other objects further require elements from First-Order Logic (at least \textit{universal} and \textit{existential quantification}). Participants were given feedback on the correct labeling of objects after every set. 

For use with language models, we reframed the word learning task as a binary classification task in which the model's job is to label whether objects are in the class described by the rule, and we converted the visual objects to text representations (e.g. `medium blue rectangle'). To create an analog of the feedback that human participants received, we provided the history of correct labels for prior object sets when querying the model to label the next object set (see Figure \ref{fig:samplegraph}).

\paragraph{Input Data} There were 112 distinct rules constructed in PTG16, with 34 rules that could be expressed using only propositional logic (``Propositional Rules") and 78 rules requiring also first-order logic (``FOL Rules"). The propositional rules involved feature primitives combined with the operators \textit{not, or, and} and \textit{xor}. The FOL rules involved existential and universal quantification over an object set, with uniqueness judgments (e.g. ``\texttt{the only blue object}") and feature comparisons (e.g. ``\texttt{same color as a triangle}" or ``\texttt{largest circle}"). As described in PTG16, rules were not exhaustive but chosen to cover a theoretically interesting range of concepts.

Each rule had an exemplar list designated in PTG16 as the held-out list, containing 25 sets of objects with labels for each object according to the rule. Each set of objects contained one to five randomly sampled objects, adding up to about 75 objects in total ($\mu$=74.7, 
$\sigma$=$6.40$) for each exemplar list. Each list was paired with human data collected by PTG16 from an average of 22 human participants per list ($\mu$=22.3, $\sigma$=1.44). Full details on exemplar list construction and human participant data collection can be found in PTG16. 

\paragraph{Models} We selected a range of recent large language models, across both chat and completion formats. Chat models are trained for a conversation format and use conversation turns as inputs and outputs, where each turn is given a speaker role attribution to the system (i.e. instructional message), the user, or the model agent; for chat models we use GPT4, GPT3.5, Llama2-70b-Chat \citep{Touvron2023Llama2O} and Mixtral 8x7B Instruct \citep{mixtralaiteamMixtralExperts2023}. Completion models take a text sequence as input and are trained to generate text that follows from the input; for completion models we use Llama2-70b and Llama2-7b \citep{Touvron2023Llama2O}, Gemma (7b) and Gemma (2b) \citep{gemmateamGemmaOpenModels2024}, GPT2 (XL) and GPT2 \citep{radfordLanguageModelsAre2019}. Although we test a variety of models, we do not aim to exhaustively test every LLM. Our main aim is to find a model that succeeds on the task, if any, and then understand how that model does so. 

We used the text output generated by the model to evaluate each model's performance. Prompting details are given in \ref{sec:promptmethod}. For completion models, objects labeled with completions that were not variants of ``True" or ``False" (subject to leading and trailing spaces and alternate capitalization) were excluded from the labeling data for these models. For chat models, labels could not be extracted for some objects because the model reiterated entirely different objects for labeling, or abstained from labeling. These objects were also excluded from the labeling data for these models. Apart from GPT2 and GPT2-XL which had an exclusion rate of 8.47\% and 3.98\% of objects respectively, all other models had an average exclusion rate of 0.058\% of objects.

\paragraph{Metrics} For each model and each rule, we have two metrics of accuracy. First, we measured the proportion of correct labels across all objects (``overall"). To account for the fact that early objects are less likely to be labeled correctly than later objects due to more limited information about the rule, we also measured 
the proportion of correct labels on just the last quarter of objects (``last-quarter"). Since every exemplar list contains about 75 objects on average, the last-quarter accuracy is computed over 18 objects on average ($\mu$=18.6, $\sigma$=1.6).

\paragraph{Human Baselines} Following the data processing procedure in PTG16, we excluded subject data that completed less than 5 sets for any concept, and also subject data that fell outside two standard deviations of accuracy scores for their attempted concept. We computed the mean human participant accuracy across rules and the corresponding propagated standard deviation bounds (i.e., the sum of variances in accuracy for each rule, divided by the number of rules). This statistic contextualizes human and model performance when aggregated across multiple rules. 

\subsection{Results}
The results are reported in Table \ref{tab:rq1results}.  Because each model can only be tested once on each exemplar list, it is not possible to sample responses from individual models or conduct standard inferential statistical tests. On propositional rules, the best models were GPT4, Mixtral 8x7b Instruct, and Gemma (7B), each of which numerically surpassed the lower standard deviation bound of mean human performance on both overall and last-quarter accuracy. On FOL rules, only Gemma (7B) surpasses this bound.

\begin{figure}[h!]
\centering
\includegraphics[width=\textwidth]{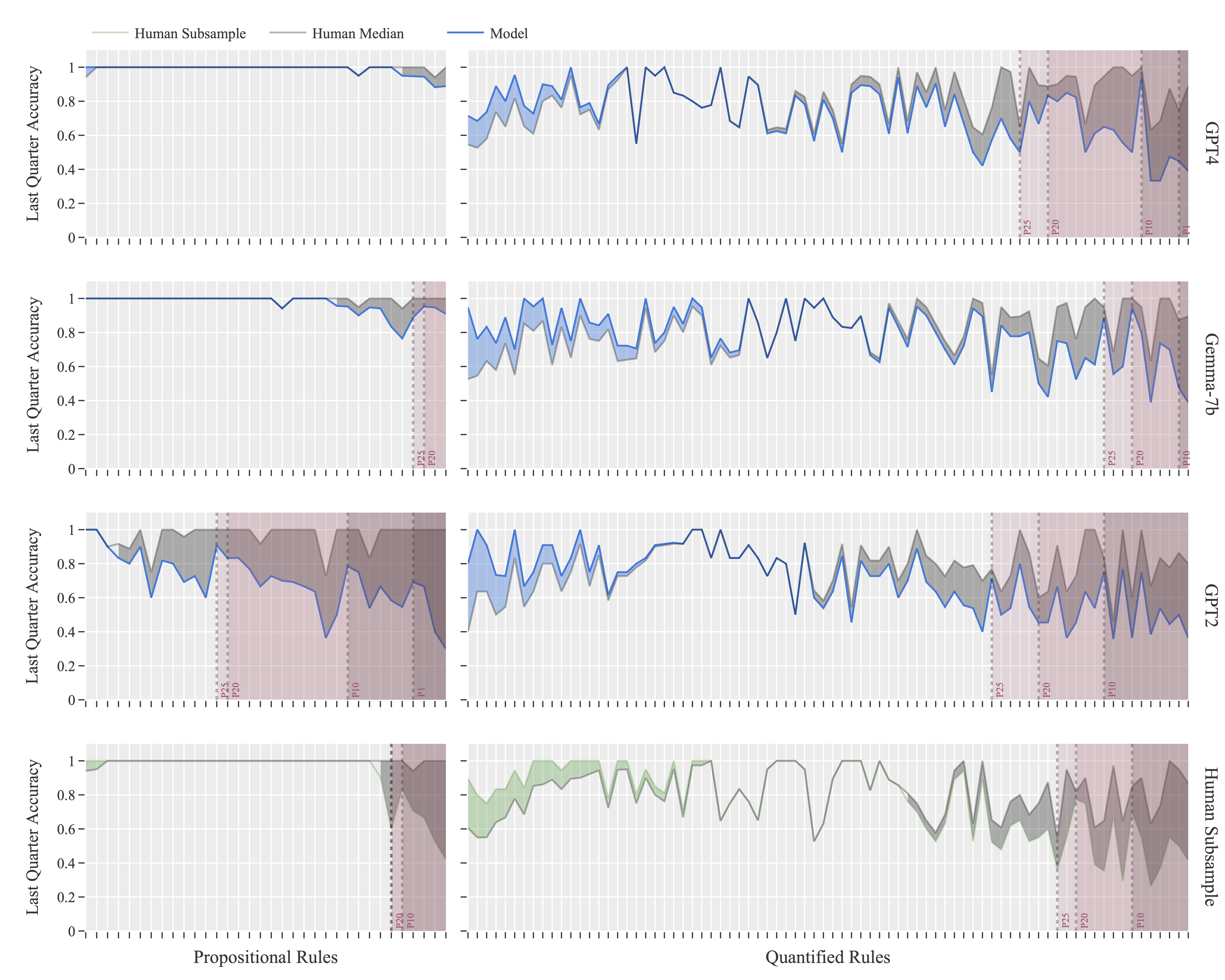}
\caption{Difference between accuracy scores between each model and the human median. Each $x$-value is a rule with two $y$-values plotted: the median last-quarter accuracy for human participants (gray) and for the model in the given row (blue). The inter-graph fill represents the difference between the last-quarter accuracies over an interval of rules: blue indicates that the model's accuracy is higher than that of the human median in that interval and gray vice versa. Within each row, rules are sorted in descending order of the accuracy difference for that model. Shaded red regions show the intervals of rules where model accuracy falls below the 25$^{th}$, 20$^{th}$, 10$^{th}$ and 1$^{st}$ percentile of human last-quarter accuracies. An analogous human subsample (green) is constructed by taking a randomly selected human participant's score for each rule. Rules are split into rule type by column.}
\label{fig:deltagraph}
\end{figure}

Mean human accuracy and its lower standard deviation bound can be a generous baseline, as the human accuracy distribution tends to be negatively skewed for many rules. Figure \ref{fig:deltagraph} examines the top models GPT4 and Gemma-7B on a rule-by-rule basis, comparing models' last-quarter accuracy to the median and distributional spread of last-quarter accuracy of human participants' data for each rule. A human subsample is provided as a qualitative human reference for how an analogous measure from human participants would deviate from the aggregate median. The subsample is constructed by first applying the same preprocessing procedure as PTG16 -- removing participants who fell more than 2 standard deviations below the mean on a given rule and those that completed fewer than 5 sets of objects -- and then randomly sampling one human participant's performance as the metric of accuracy for each rule. The worst-performing model, GPT2, is also provided as a baseline.

On a rule-by-rule basis, the top models remain within the top three quartiles of human-level accuracies for a majority of rules, and are otherwise comparable with human subsamples in the percentage of rules where they fall in the first quartile. For propositional rules, the top models GPT4 and Gemma (7B) remain within the top three quartiles of human accuracies for 100\% and 88.2\% of rules, respectively.  For FOL rules, GPT4 and Gemma (7B) remain within the top three quartiles of human accuracy for 75.6\% and 87.2\% of rules. Across all rules, the number of rules for which models fall in the bottom quartile of human performance is 17.0\% for GPT4 and 12.5\% for Gemma (7B), which is comparable to human subsamples ($\mu$=0.183, $\sigma=0.036$ from 10\,000 subsamples).

\begin{center}
\begin{table}[!h]
  \scriptsize
  \centering
  \begin{tabular}{p{9em}|C{3.8em}|C{3.8em}|C{3.8em}|C{3.8em}|C{3.8em}|C{3.8em}}
  \bottomrule
  & \multicolumn{2}{c|}{\textbf{All Rules}} & \multicolumn{2}{c|}{\textbf{Propositional Rules}} & \multicolumn{2}{c}{\textbf{FOL Rules}} \\
  \toprule 
  Model & Overall & Last Quarter & Overall & Last Quarter & Overall & Last Quarter \\
  \midrule
  \textit{Chat Models}  & & & & &\\
   GPT4 & \underline{0.764} & 0.813 & \textbf{\underline{0.908}} & \textbf{\underline{0.987}} & 0.701 & 0.737 \\
GPT3.5 & 0.685 & 0.746 & \underline{0.761} & 0.857 & 0.652 & 0.697 \\
Llama2-70b-Chat & 0.681 & 0.716 & \underline{0.787} & 0.829 & 0.635 & 0.666 \\
Mixtral 8x7B-It & \underline{0.769} & 0.815 & \underline{0.875} & \underline{0.953} & 0.723 & 0.754 \\
& & & & & \\
\textit{Completion Models} & & & & & \\
Llama2-70b & 0.734 & 0.794 & \underline{0.799} & \underline{0.916} & 0.706 & 0.740 \\
Llama2-7b & 0.760 & 0.799 & \underline{0.820} & 0.892 & \underline{0.734} & 0.758 \\
Gemma-7b & \textbf{\underline{0.795}} & \textbf{\underline{0.840}} & \underline{0.886} & \underline{0.969} & \textbf{\underline{0.756}} & \textbf{\underline{0.784}} \\
Gemma-2b & 0.751 & 0.772 & \underline{0.771} & 0.812 & \underline{0.742} & 0.754 \\
GPT2-XL* & 0.654 & 0.684 & 0.663 & 0.722 & 0.650 & 0.667 \\
GPT2* & 0.622 & 0.707 & 0.634 & 0.703 & 0.617 & 0.709 \\
\midrule
Human (25 Sets) & $0.774 \pm 0.0109$ & $0.835 \pm 0.0148$ & $0.858 \pm 0.106$ & $0.932 \pm 0.0235$ & $0.737 \pm 0.0132$ & $0.792 \pm 0.0187$\\
Human* (14 Sets) & $0.726 \pm 0.0110$ & $0.798 \pm 0.0170$ & $0.798 \pm 0.020$ & $0.900 \pm 0.028$ & $0.695 \pm 0.0132$ & $0.753 \pm 0.021$\\
  \bottomrule
  \end{tabular}
  \caption{Results for all rules (left) and for the two subsets of rules requiring only propositional logic (middle) and those also requiring first-order-logic (FOL; right) to formulate. Model averages that are above the lower SD bound of human averages for the corresponding type of rule are underlined. The highest scores for each type of rule are bolded. (*=We evaluate the GPT2 models only up to 14 object sets due to their limited context length, and accordingly compare them to human baselines on 14 sets.)}
\label{tab:rq1results}
\end{table}
\end{center}

\subsection{Discussion}
The results of this experiment demonstrate that some models succeed on the logical rule-learning task at rates comparable to human participants. 
This success requires models to demonstrate two capabilities. First, models must infer a complex labeling function from exemplars; and second, they must then apply the rule correctly to a new set of objects. 

For GPT4 specifically, these results run counter to findings in \citet{ellisHumanlikeFewShotLearning2023} that reported GPT4 as well-below human accuracy when asked to perform this same task with the same stimuli. This difference may have been caused by differences in prompts, or by closed-source updates made to GPT-4 (see \ref{sec:elliscomparison}). In general, our choices in textual encoding of the task may have influenced the models' final performance, and alternative representations (e.g. a JSON encoding) may have yielded different absolute numbers. However, since we are not optimizing for the \textit{best} model performance, our naive textual encoding has sufficed to provide an affirmative answer to the question of whether any LLM could be comparable in performance to human learners in this rule learning task.

Nevertheless, high classification accuracy alone is not sufficient to establish that the models have induced logically structured rules, let alone the same rules humans induce in this task. These models have pretrained representations of all of the color, size, and shape words used to describe object features, and they may rely on approximate, non-human-like heuristics (see, e.g. \citep{mccoyRightWrongReasons2019}). They may also compute complex, ineffable functions that approximate the correct classification of exemplars with these specific features into `True' and `False', without deploying any logic-like operators whatsoever. It is also possible that even if the models do use logical operators, they cannot articulate their rules in logical terms, just as humans can implicitly learn rules that they cannot subsequently articulate \citep{reber1989implicit}. The high accuracy of these models in Experiment 1 enables us to ask to what degree these models can articulate logical elements of the rules that they are employing.  We use one of the top-performing models, GPT4, to approach this question in Experiment 2.

\section{Experiment 2: Rule Elicitation}
In this experiment, we prompt GPT4 with a sequence of object sets to induce a rule, and then ask it both to describe the classification rule it has inferred, and also classify objects in an unlabeled object set. We choose GPT4 because it was both a top-performing model in Experiment 1 and, due to its tuning to provide helpful replies in conversation, is able to readily produce verbal explanations in its off-the-shelf state.

There are known limitations to interpreting LLMs' self-reported explanations of their own behavior. Because LLMs predict the most likely next word(s) given the words in their input prompt, nothing guarantees that an LLM's self-reported explanation of its behavior is connected in any way to the actual process that generated that behavior. In fact, prior work has shown that models can produce plausible explanations to justify ad-hoc any answer that initial system messages instruct them to choose \citep{turpinLanguageModelsDon2023}. It is therefore possible that the model could report a different rule than the one it is actually using to classify. 

To test for the correspondence between the model's classification and its self-reported rule, we test whether applying that rule would actually produce that classification of objects at each timestep. We can then also ask whether the model's self-reported rule matches the correct rule (that is, the rule that was actually used to determine whether the object is `wudsy' in PTG16). We do not, however, directly compare the rules that models and humans report, as we do not have data on the rules humans (purport to) use. Although PTG16 asked their human participants to describe the rules they were applying, they reported that these descriptions were generally much less specific than the actual rule participants seemed to be applying, and did not report or analyze those results.

\subsection{Method}
\paragraph{Prompting} In each prompt to GPT4, we included a system message of instructions, sets of labeled exemplars, and then the new object set that was to be labeled. The system message instructs GPT4 to infer a labeling rule from the labeled exemplars, explain the labeling rule concisely, and then provide labels for the new object set ( \ref{sec:promptmethod}). This procedure was iterated for every subsequent object set, following the same lists used in PTG16 and Experiment 1.

\paragraph{Rule Evaluation} GPT4 generated one self-reported rule per object set, for a total of 850 generated rule descriptions for the 34 Propositional rules, and 1950 generated rule descriptions for the 78 First-Order Logic (FOL) rules. To evaluate whether the self-reported rule matches the classification, we converted the natural language rule to Python code. We used GPT 3.5 to produce a first-cut conversion, then manually reviewed all rule descriptions and revised code conversions for all propositional rules and a random sample of 16 FOL rules (i.e. 20\% of FOL rules). We also reviewed the final rule descriptions generated at the last, `unobserved' object sets for all rules.

\paragraph{Metrics}

In addition to qualitative analysis of the elicited rules and the same last-quarter accuracy measure we used in Experiment 1, we also report three aggregate quantitative metrics: likelihood, match rate and consistency. \textit{Likelihood} is defined as the percentage of labels in previously seen sets that each hypothesized rule correctly accounts for, indicating the degree to which the model's described rules are consistent with the provided evidence. \textit{Match rate} is the proportion of rules where the model's reported rule for the final, unobserved set of exemplars is truth-functionally equivalent to the true underlying rule, defined as whether the final rule has a likelihood of 1.0. \textit{Consistency} is the percentage of objects labeled by the model on which the model's classification aligns with the rule it reports using. We take high consistency to be evidence that the model employs its self-reported rule for its classification behavior, or more precisely, that both of these reports have a common underlying representation.

\paragraph{Bayesian pLoT Baseline} While PTG16 suggest that it is not possible to elicit rules from human participants, it is possible to compare GPT4's rules to the Maximum A Posteriori (MAP) rule for each object set generated by a Bayesian pLoT model that fits human participants' data well. We extract only the MAP rule, under the assumption that since it contributes the most probability mass to the posterior of a classification decision, it can be interpreted as the Bayesian pLoT model's inferred rule for the task to a first approximation.

We use the pLoT Bayesian model data reported in \citet{ellisHumanlikeFewShotLearning2023}'s Bayesian Program Learning baseline, which was PTG16's model re-fit via the same procedure over the same data using the best-fitting FOL grammar from PTG16's paper. These object lists are the same as those given to GPT4. We manually converted these MAP rules, which are natively lambda expressions, into Python code to use them to classify objects. We considered the pLoT Bayesian's model's label to be `True' for an object if the posterior probability assigned to the `True' label for the object is greater than 0.5.

\subsection{Results}
\paragraph{Propositional Rules}  With rule elicitation included in the prompt, the LLM's mean last-quarter accuracy on the classification task was 0.934. This was a slight decrease from its last-quarter accuracy without rule elicitation in Experiment 1 (0.987), but is still comparable to the accuracy of human participants ($0.932 \pm 0.0235$). Meanwhile, the Bayesian model attained a mean last-quarter accuracy of 0.994. The LLM's classification responses were consistent with its self-reported rule for 96.3\% of object labels, and its proposed rules across all object sets had an average likelihood of 90.1\%. In comparison, the Bayesian pLoT model's MAP rules had a consistency of 90.2\% and an average likelihood of 95.7\%. The imperfect consistency for the Bayesian pLoT model is the result of its noise parameter $\alpha$, where PTG16 defined the model with a $(1-\alpha)$ chance of not labeling according to the rule, but instead according to a baseline distribution where the `True' label has $\beta$ probability. The values for $\alpha$ and $\beta$ were obtained from fitting to human participants' data, with full fitting details reported in PTG16.

The match rate of the LLM was much lower than that of the Bayesian model. The LLM arrived at a truth-conditional equivalent of the underlying rule for 44.1\% of rules at the last object set, while the Bayesian model did so for 82.4\% of rules. These rates are, however, difficult to interpret without knowing the rate at which human participants themselves would report a truth-conditional equivalent of the underlying rule. While PTG16 suggest that obtaining humans' own reports may not be possible because humans may not represent these rules either in natural language or in another effable form, it is also possible that future work could elicit interpretable rule reports from humans using different prompts or higher incentives.

Table \ref{tab:sample-rules} shows a representative sample of the rules that the models produced at the final, `unobserved' object set. The LLM's generated rules mostly involve combinations of only \texttt{`and'}, \texttt{`or'} and \texttt{`not'} operators, and tend to concatenate many more features together than the Bayesian model's MAP rules. Rules (1-2) show that the model is able to express \texttt{`not'} rules through their complements, and Rules (3-5) show that it is able to express operators like \texttt{`implies'} through complex combinations of \texttt{`and'}, \texttt{`or'} and \texttt{`not'}.  Even when it does not recover the rule exactly, Rule (5) shows how the LLM can still offer an approximation that has a high likelihood -- which in this case is equivalent to the MAP rule. The LLM demonstrates a wide variance in its performance even where the true rules have similar complexity; Rules 6-10 all combine two primitives with either an \texttt{`and'} or \texttt{`or'} operator, but the LLM demonstrates a lot of variance in its ability to recover these rules.
The LLM also clearly struggled to learn even a good approximation for any of the three \texttt{`xor'} rules in (11-13), where the Bayesian pLoT model is easily able to use its \texttt{`iff'} primitive to express this concept. These results raise an important open question for further research on humans: do the rules people induce actually include more complex logical primitives, such as \texttt{`xor'} and \texttt{`iff'}, or do they only approximate these rules with simpler operators?  

\begin{table}
\scriptsize
\centering
\renewcommand{\arraystretch}{1.7}
\begin{tabular}{l||p{10em}||p{12em}|c||p{8em}|c}
\# & \textbf{True Rule} & \textbf{LLM Rule} & \textbf{$\mathcal{L}$} & \textbf{MAP Rule} & \textbf{$\mathcal{L}$}\\
\toprule
1 & not circle & \textbf{triangle or rectangle} & 1.00 & \textbf{not circle} & 1.00\\
2 & circle and (not blue)& \textbf{(green or yellow) circle} & 1.00 & green circle & 0.959\\
\midrule
3 & (not blue) implies (not circle) & \textbf{blue or (green and not circle) or yellow (rectangle or triangle)} & 1.00 & \textbf{(not circle) or blue} & 1.00\\
4 & circle or (triangle implies blue) & \textbf{rectangle or circle or (blue triangle)} & 1.00 & \textbf{(not triangle) or blue} & 1.00\\
5 & circle implies blue & rectangle or triangle & 0.93 & not circle & 0.93 \\
\midrule
6 & circle or blue & \textbf{blue or (yellow or green circle)} & 1.00 & \textbf{circle or blue} & 1.00\\
7 & blue or green & green or ((blue (triangle or rectangle)) & 0.95 & \textbf{not yellow} & 1.00\\
8 & blue or small & small (triangle or rectangle)	& 0.70 & \textbf{blue or small} & 1.00\\
\midrule
9 & small and blue & \textbf{small blue (circle or rectangle or triangle)} & 1.00 & \textbf{small and blue} & 1.00 \\
10 & circle and blue & large and blue & 0.85 & larger than itself & 0.91 \\
\midrule
11 & circle xor (not blue) & not (medium blue) and not (large blue) & 0.74 & \textbf{blue iff circle} & 1.00\\
12 & circle xor blue & ((large green) or (medium yellow)) circle & 0.71 & \textbf{circle iff (yellow or green)} & 1.00\\
13 & not (circle xor blue)	& green or (triangle) & 0.53 & \textbf{blue iff circle} & 1.00
\end{tabular}
\caption{Examples of the model's described rule at the final set compared to the true rule, for where the true rule is propositional. Likelihood ($\mathcal{L}$) describes the proportion of known labels that the rule correctly accounts for; model rules that are truth-conditionally equivalent to the true rule are bolded.}
\label{tab:sample-rules}
\end{table}

\paragraph{FOL Rules} As in Experiment 1, GPT4's performance on rules that require first-order logic to represent was significantly lower than on propositional rules. Prompted with rule elicitation, the LLM classified objects with a mean last-quarter accuracy of 0.687. This is a further decrease from its accuracy in Experiment 1, in which the LLM was not prompted for rule elicitation (0.737). Human last-quarter accuracy on FOL rules was higher ($0.792 \pm 0.0187$), and that of a Bayesian pLoT model even higher at 0.881.
 
On an annotated random subset of 20\% of FOL rules, the LLM showed a consistency of 95.5\% between its self-reported rule and its classification, and mean likelihood of 76.6\% of generating a rule consistent with prior evidence. The Bayesian pLoT model's MAP rules on the same subset have a lower consistency of 89.64\%, but a higher mean likelihood of 88.9\%. Examining the last object sets of all rules, the LLM does not arrive at a truth-conditional equivalent for any rule, while the Bayesian pLoT model does so for 22.7\% of rules. Notably, the LLM does not invoke first-order logic for any of its final rules at the last object set. There are some object sets in the annotated subset where the LLM does verbalize a rule using first-order logic (FOL), but only for a minority (4.4\%) of the annotated subset. None of these match the right FOL rule. Some examples are shown in \ref{sec:fol-attempts}.

Table \ref{tab:sample-rules-fol} shows a representative sample of rules that either model reported at the last set of objects. In the previous results for propositional rules, LLM rules tended to be verbose,  stringing together disparate features. This tendency is even more pronounced for FOL rules, where the model appears to be stringing together features of previously positive exemplars. Sometimes, this strategy lends itself to a good approximation of the underlying rule that is comparable to the approximation that the MAP rule produces (e.g. Rules 1 to 6). However, there are more cases where the strategy achieves a poor likelihood, while the Bayesian pLoT model recovers the true rule (e.g. Rules 7 to 9).

\begin{table}
\scriptsize
\centering
\renewcommand{\arraystretch}{1.7}
\begin{tabular}{l||p{9em}||p{12em}|c||p{10em}|c}
\# & \textbf{True Rule} & \textbf{LLM Rule} & \textbf{$\mathcal{L}$} & \textbf{MAP Rule} & \textbf{$\mathcal{L}$}\\
\toprule
1 & blue or (not smaller than any other object) & large or blue & 0.91 & large or blue & 0.91\\
2 & the unique object that is a blue circle & small blue circle & 0.98 & larger than itself & 0.97 \\
3 & unique blue object & small blue (rectangle or triangle) & 0.82 & larger than itself & 0.86 \\
4 & same shape as one of the largest and blue & blue (circle or rectangle) & 0.85 & (not green) and (not yellow) & 0.85 \\
5 & there is a triangle in the set & not (small or blue) or (small (yellow triangle or blue circle)) & 0.68 & triangle & 0.57 \\
6 & larger than a blue object & large and (blue or yellow) & 0.78 &  \textbf{larger than a blue object} & 1.0 \\
7 & same color as all other objects of the same shape & blue or (triangle) or (green circle) or (yellow rectangle) & 0.58 & \textbf{no object has the same shape but not the same color} & 1.0 \\
8 & one of the smallest & large or (medium triangle) or (small (rectangle or triangle)) & 0.35 & \textbf{not larger than any object in the set} & 1.0 \\
9 & same color as another object & large circle or medium triangle or (small circle and (not green)) & 0.39 & \textbf{same color as an object that is not the same size or not the same shape} & 1.0\\
\midrule
\end{tabular}
\caption{Examples of the model's described rule at the final set compared to the true rule, for where the true rule requires first-order logic. Likelihood ($\mathcal{L}$) describes the proportion of known labels that the rule correctly accounts for; model rules that are truth-conditionally equivalent to the true rule are bolded.}
\label{tab:sample-rules-fol}
\end{table}

\subsection{Discussion}
We find that GPT4 is able to report, with very high consistency, the rules that it actually appears to use to classify objects. Since this LLM is pretrained to generate text that is coherent with the prompted context, and further tuned with human feedback to provide answers relevant to the user's queries, it is unsurprising that the model would express \textit{some} rule in the correct domain space. It is not surprising, for instance, that the elicited rules use words for logical operators or only those in-domain features mentioned in the prompt. However, there was no guarantee that any of these elicited rules would be strongly consistent with the model's classification, or that they would have a high likelihood of accounting for known evidence (as was the case especially with propositional rules). Thus, the most parsimonious interpretation of high consistency is that the model's stated rules and its classification behavior stem from a common representation.

At the same time, examining these self-reported rules highlights that just because different learners (human, Bayesian pLoT model, and LLM) attain comparable accuracy on the task, they do not necessarily solve the task in the same way. %
For propositional rules, we find that the LLM achieves human-level classification success, but that its self-reported rules tend to be more verbose than the Bayesian MAP alternative, albeit without any loss in likelihood. Thus, based on accuracy metrics alone and lacking interpretable reports from humans about the rules that they are using \citep{piantadosiLogicalPrimitivesThought2016}, it is possible that humans are inferring rules that are more like the longer rules of GPT4 than the shorter rules of the Bayesian pLoT model. Taking together the present results with PTG16, it is possible that unlike the Bayesian model, humans do not actually have a strong bias for shorter rules \citep[cf.][]{piantadosiLogicalPrimitivesThought2016,feldmanSimplicityPrincipleHuman2003}, but a quite different tendency to concatenate additional representational elements to accommodate new exemplars. 

For FOL rules, we find that the LLM fails to use first-order logic, and instead attempts to approximate the evidence via complex chains of propositional operators. Although the LLM's accuracy with rule elicitation is lower than that of humans on FOL rules, its accuracy on the same rules in Experiment 1 (when it was not asked to report on the rule it inferred) was higher and comparable to human participants. At the same time, the pLoT model we chose for comparison was the best-performing model of those that included FOL primitives from PTG16. As PTG16 report, pLoT models with FOL primitives were a substantially \textit{worse} match to human learning trajectories on FOL rules than what pLoT models with Boolean operators could achieve on Boolean rules, leading PTG16 to suggest that human quantificational reasoning may not be captured by a model that includes the quantifiers as primitives.

Our findings suggest another possibility: that humans, like LLMs, may in some cases formulate rules that match the ground truth rule better in their extension (i.e. which objects are 'wudsy') than their intension (i.e. being formulated not over quantifiers, but over more complex compositions of propositional operators). This may have less to do with the capacity to represent quantifiers as primitives, and more to do with people's assumptions about the task: that an object-classification task would not be designed with rules in which whether one object in the set is `wudsy' depends on whether that object is "the same color as all other objects of the same shape" (Table 3, Rule 7). At the same time, some rules with higher-order primitives may be more accessible to humans than to LLMs in this task (e.g. Rule 9: "same color as another object"; see also, Figure 4). It is possible that if quantifier words were explicitly listed in the model prompt, the LLM would be more likely to use those words in its elicited rules. However, this would be a different distribution over possible hypotheses than that obtained from general language prediction over a large corpus, which has been shown to be an accurate prior for human hypothesis generation in this task \citep{ellisHumanlikeFewShotLearning2023}. We do not explore this possibility here, though it may be an interesting direction for future work.

Given that overall accuracy metrics alone leave the task strategy underdetermined, we use a more fine-grained metric of $R^2$ with individual human responses in Experiment 3. Here, a high $R^2$ between any pair of model and human responses requires that the model produce labeling behavior that matches the successes and mistakes that human participants demonstrate when learning a rule object by object. For $R^2$ to be high, the model must attain a matching learning trajectory across the rule, over the sequence of objects. For that to obtain, not only would the hypotheses that the model generates need to be similar, but also the search space from which the model iteratively induces hypotheses and the procedure for selecting hypotheses given evidence.
 
\section{Experiment 3: Correlation with Human Learning Trajectories}
In this experiment, we examine if a model that succeeds on the rule learning task performs the task in a human-like way, as quantified by $R^2$ with human responses, object-by-object. Succeeding on the task does not trivially entail a high correlation with human behavior. While GPT4 also demonstrates strong performance in Experiment 1, \citet{ellisHumanlikeFewShotLearning2023} reports that its $R^2$ with human responses is only around $0.42$. We select Gemma (7B) to use in this experiment due to the combination of its competitive performance in Experiment 1 and the availability of its weights. The open weights enable calibration of model parameters to human response data, which allows for a fairer comparison to the Bayesian model that has undergone parameter-tuning on the same human data.

\subsection{Method}
\paragraph{Metrics} We measure the probability of the same label being assigned by human participants and the LLM on each exemplar of each rule, and then compute the $R^2$ (square of Pearson's correlation coefficient) between humans' and LLMs' classification across all rules and objects. For the LLM, this is the model's probability assignment to the `True' label for each object. For human participants, this is the proportion of human participants that labeled the object as \textit{wudsy}.  If the proportion of human participants assigning this label is interpreted as an approximation of the underlying human posterior probability of that classification, then a high $R^2$ indicates that the model's probability assignments track human posterior probabilities well at an object-by-object granularity. 

\paragraph{Finetuning} PTG16's Bayesian models included model parameters that were fitted to human data. For a fairer comparison between these models and the LLM, we fine-tune the LLM to match its output probabilities to a training set of human data, before testing on a different set of withheld data. 
The target distribution for each exemplar is generated by assigning the proportion of human participants that responded \textit{wudsy} as the probability of the ``True" token,  and the complement as the probability of the ``False" token, with zero elsewhere for all other tokens in the model's vocabulary. We fine-tune by minimizing cross-entropy loss between the target and model distributions (Equation \ref{eq:loss}).

\begin{equation}
\mathcal{L} = - \sum_x P_x(t) \ln Q_x(t) = \sum_x \sum_{t\in\{\text{``True"}, \text{``False"}\}} P_x(t) \ln Q_x(t)
\label{eq:loss}
\end{equation}

\noindent where $x$ are objects to label and $t$ is a token in the model's vocabulary. $P_x(t)$ and $Q_x(t)$ are the target and model distributions' probability assignments for token $t$ at object $x$. Since $P_x(t)$ is zero everywhere except $t\in\{\text{``True"}, \text{``False"}\}$, the loss term needs only to sum over these two tokens for each $x$. 

We compare the results of this set up with a fine-tuning ablation that minimizes the cross-entropy loss between the models' outputted labels for each object and their true labels, i.e. tuning to the true labels of training lists instead of tuning to human responses. All finetuning was done on two GPUs using QLoRA \citep{dettmers2023qlora}, tuning about 2\% of model parameters with the model quantized at 4-bit. Details are in \ref{sec:ftmethod}.

\paragraph{Data} We use exactly the training and held-out exemplar lists for each rule as designated in PTG16.  We tune on the designated training lists, while reporting $R^2$ metrics computed only on held-out lists that are never seen by the model. This is identical to the testing setup for PTG16's Bayesian pLoT models. 

\subsection{Results}
\begin{figure}
    \includegraphics[width=\linewidth]{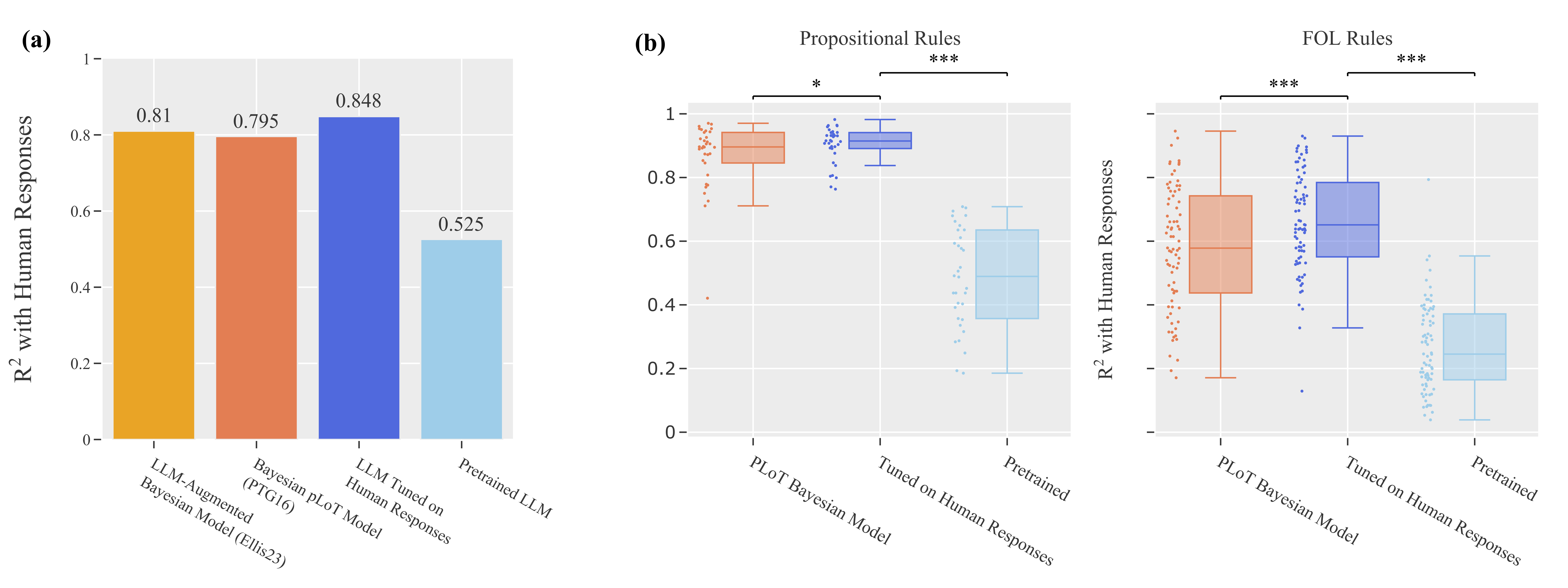}
    \caption{$R^2$ values between model and human posterior probabilities for responding \textit{True} between different models. (a) $R^2$ values across all objects from all rules. (b) $R^2$ values plotted where each scatterpoint is a rule. Significance testing was done using two-tailed paired t-tests.} \label{fig:r2a}
\end{figure}

\begin{figure}[t!]
    \centering
    \includegraphics[width=\linewidth]{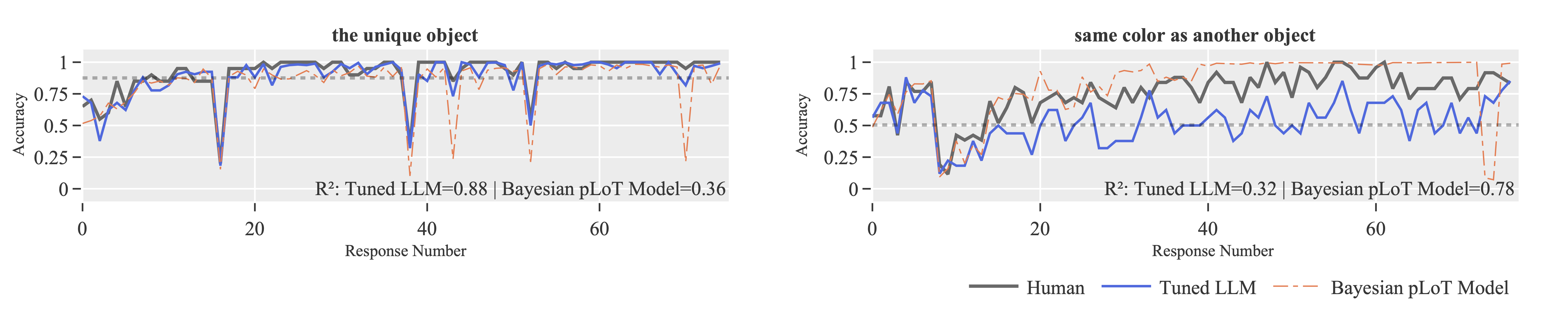}
  \caption{Learning trajectories compared between human participants, the tuned LLM, and the pLoT model with a FOL grammar from PTG16. The two rules were chosen to be representative of a good and bad fit for the Tuned LLM compared to the pLoT model. The gray dashed line shows the chance baseline, defined as the accuracy from guessing `True' at the empirical rate of `True' labels for the exemplar list.} \label{fig:trajectories-w-ptg16}
\end{figure}

Figure \ref{fig:r2a}(a) shows the $R^2$ between different LLM variants' probability assignments to `True' with that of human participants. We compare them to the $R^2$ reported in the LLM-augmented Bayesian model from \citet{ellisHumanlikeFewShotLearning2023}, and the $R^2$ computed using the posterior probabilities generated from PTG16's pLoT Bayesian model using a FOL grammar. We note that the $R^2$ value for the pLoT Bayesian model is not the same as that reported in PTG16, as we manually re-compute the value using the prediction data generated from the pLoT Bayesian model reported in \citet{ellisHumanlikeFewShotLearning2023}'s Bayesian Program Learning baseline. This prediction data was generated using PTG16's Bayesian pLoT model design with the best-performing FOL grammar from PTG16, re-fit via the same procedure on the same data, but ultimately may be slightly different than the original model from which the $R^2$ in PTG16's paper was derived due to stochasticity in the learning procedure.  

Once tuned on training lists of human responses, as is also done for both Ellis's and PTG's Bayesian models, the LLM explains 84.8\% of variance in human participant responses in held-out lists. 
As Figure \ref{fig:r2a}(b) shows, on a rule-by-rule basis, 
the learning trajectory that the tuned model demonstrates has a higher correlation to human responses than the Bayesian pLoT model.

Two learning trajectories produced by the tuned LLM and the Bayesian pLoT model are shown in Figure \ref{fig:trajectories-w-ptg16}, representative of a typical good and bad fit for the tuned LLM. In cases of good fits, as achieved on \textit{`the unique object'}, the tuned LLM closely matches the occurrences of peaks and troughs in human learning trajectories, and qualitatively seems to match the magnitude of these changes better than the pLoT model. 
PTG16 interpreted steep troughs in human learning curves, especially those that the Bayesian pLoT model also reproduced, as indicating exemplars that falsified the best hypothesis about the rule up to that point. Following this interpretation, the fact that the tuned LLM frequently matched the occurrences of these troughs strongly suggests not only that it often arrived at a similar best-so-far hypothesis, but also that it may be implementing a similar inference procedure as both human participants and the Bayesian pLoT model. On the other hand, on the minority of rules where the tuned LLM has a much lower $R^2$ with humans than the Bayesian model, such as in the bad fit observed on `\textit{same color as another object}', the tuned LLM failed to learn the rule at all. The tuned LLM hovers around the chance baseline while human participants and the pLoT Bayesian model attain high accuracy by the late exemplars. \ref{sec:additional-traj} provides additional examples of rule-learning trajectories, illustrating these patterns for other rules that the LLM fits well and those it fits poorly.

\begin{figure}[h!]
    \centering
    \includegraphics[width=\linewidth]{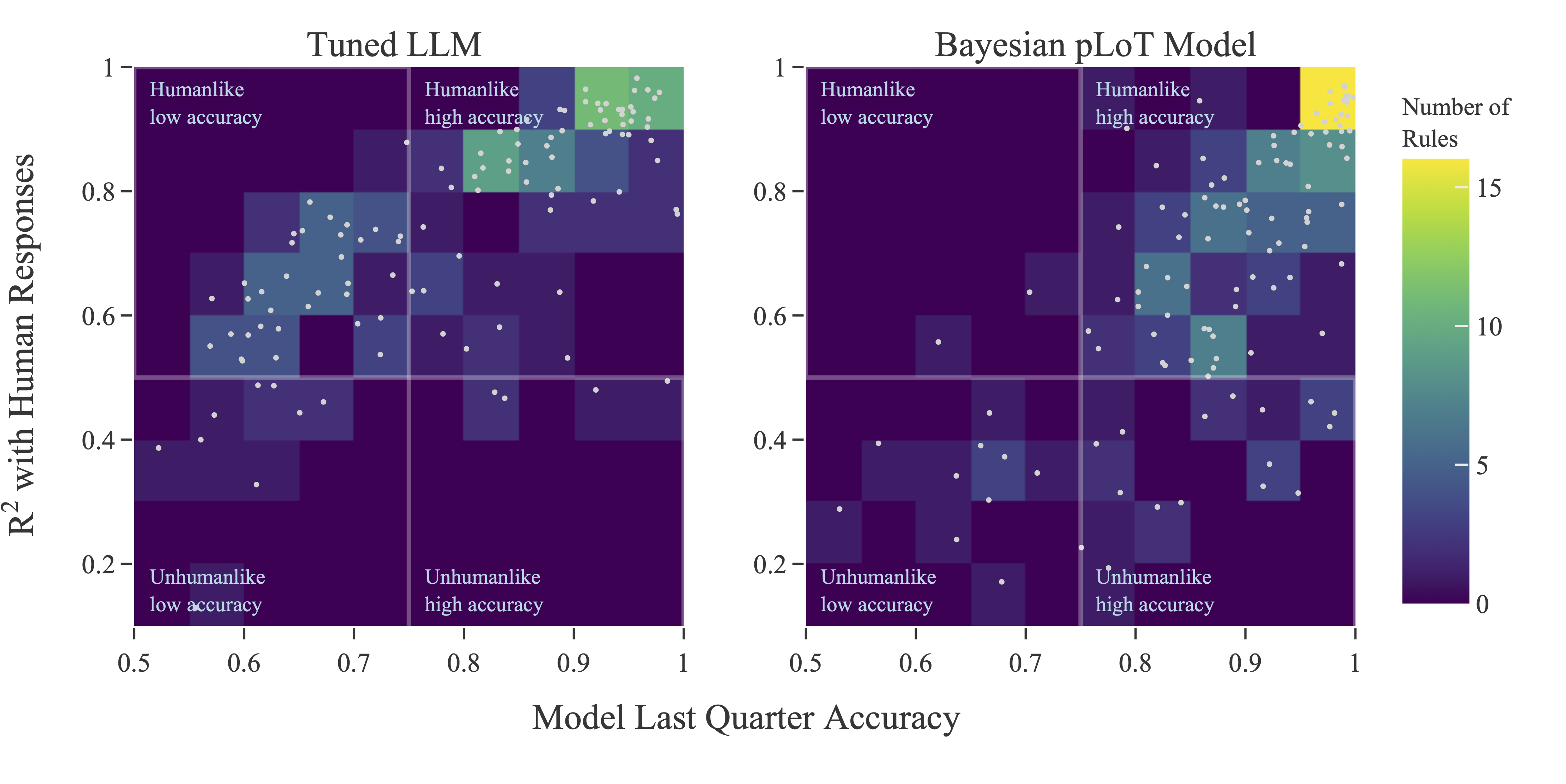}  
    \caption{How each model's $R^2$ with human responses changes with the model's last-quarter accuracy on each rule. Quadrant annotations are guides for qualitative interpretation. First, the LLM has overall higher $R^2$ with humans. Second, on rules where the tuned model achieves high accuracy, it tends to also have a high $R^2$ with human responses; when the Bayesian model achieves high accuracy, it may still have a low $R^2$ with humans.} \label{fig:vsbayesianscatter}
\end{figure}

An aggregate view over all rules in Figure \ref{fig:vsbayesianscatter} supports these qualitative observations. When the tuned model achieves high accuracy on a rule, it tends also to demonstrate a high correlation with human responses. Conversely, when the Bayesian pLoT model succeeds on a rule, its learning trajectory does not necessarily correlate highly with human responses. We tested this difference between the Bayesian model and the LLM by fitting a linear mixed-effect model to predict $R^2$ with human responses. We fit the predictors of last-quarter accuracy, a categorical variable for the model (LLM or Bayesian pLoT), and their interaction, with a random intercept for individual rules, and $p$-values generated from Type-II Wald Chi-Square tests. We find significant main effects of accuracy ($\chi^2$ = 129.29, \textit{p} $< 0.001$), and of the model ($\chi^2$ = 101.21, \textit{p} $< 0.001$), confirming the LLM's better fit to human data. Importantly, we also find a highly significant interaction term  $(\chi^2$ = 13.74, \textit{p} $< 0.001)$, indicating that while there are many non-human-like ways both models can fail to learn a rule, the tuned LLM is more likely to succeed exclusively in a human-like way. 

\begin{figure}[h!]
    \centering
    \includegraphics[width=\linewidth]{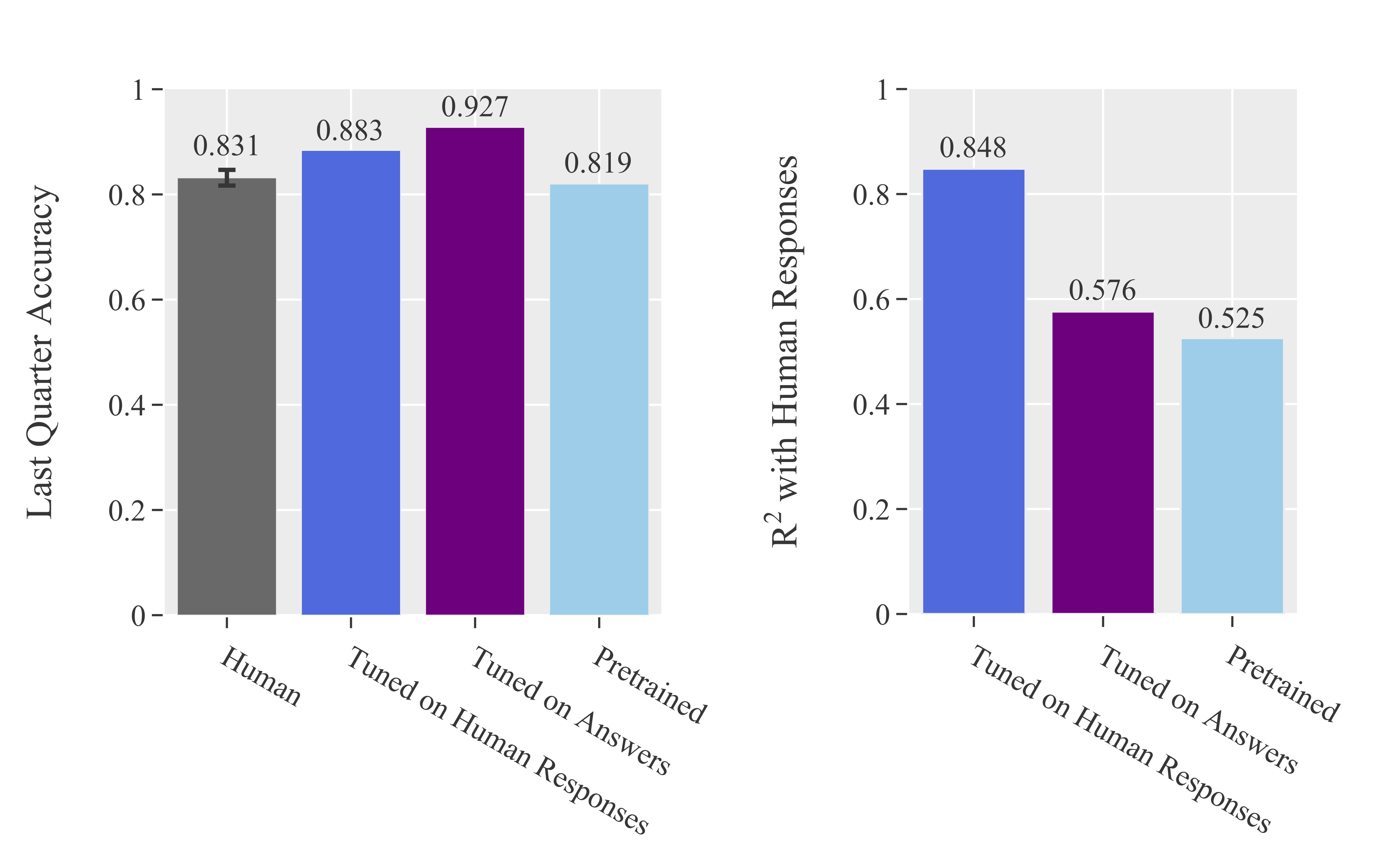}
    \caption{Comparison of the relationship between accuracy and $R^2$ between the Bayesian model and tuned LLM. The LLM tuned on answers attains a higher last-quarter accuracy on the task than that tuned on human responses, but it has a poor correlation with human responses. Its improvement in correlation over the pretrained model can be interpreted as the correlation accounted for by learning rules correctly.} \label{fig:accuracyvsr2}
\end{figure}

Figure \ref{fig:accuracyvsr2} compares the $R^2$ values and accuracy scores between the LLM tuned on human responses and the LLM tuned on the correct labels (from the fine-tuning ablation that used cross-entropy loss over the correct labels). The model tuned on the correct labels achieves a high accuracy of 0.927, but only explains 57.6\% of variance in human responses. Since this $R^2$ value is still higher than the pretrained model, it may reflect the increase of fit to human responses that is necessarily entailed just by learning rules successfully.
The improved fit—and lower accuracy—of the LLM tuned on human responses demonstrates that its high correlation with human rule-learning trajectories is the result of more than simply learning the rules correctly. 

\subsection{Discussion}

When given the opportunity to tune to human responses, as PTG16 had done for their Bayesian models, the tuned LLM fits human learning trajectories significantly better than the best Bayesian pLoT models extant in the literature. Not only that, but when the LLM successfully learned a rule, it tended to do so in a human-like way. These results may be especially surprising considering the LLM's design.

It is not trivial that an LLM tuned to fit human behavior on training data would accurately predict human behavior in held-out data to such a degree. Given that the LLM's pretraining objective was primarily predicting the correct next word token, and that fine-tuning was performed only on a small fraction of its weights and with a small set of data compared to its pretraining corpus, it is surprising that its probability assignments to answer tokens could be calibrated to mirror human responses at all. And yet, we found that the tuned LLMs' probability assignments nonetheless did generate learning trajectories with the same peaks and troughs at similar magnitudes as were found in human participants' average learning trajectories. 

When Bayesian pLoT models achieved a high $R^2$ with held-out human learning trajectories, they were interpreted as encoding an accurate computational theory of human inference. Hence in a similar vein, our results here support the conclusion that tuned LLMs are likewise implementing similar computations as those employed by human learners.  Unlike Bayesian pLoT models, however, the tuned LLM reproduced humans' responses without any explicit inference procedure or logical primitives being built in.

It is hard to know what the LLM learns when tuning to human responses, and it is hard to fully eliminate the concern that the tuning process is introducing an artifact that increases fit even on withheld exemplars. We tried to minimize this possibility in Experiment 3: the tuning data consisted purely of exemplar lists and did not explicitly mention any logical operators. Furthermore, the difference in $R^2$ fit to humans between the LLM tuned on the correct answers and the LLM tuned on human responses suggests that just learning to successfully infer rules is insufficient to yield so strong a correlation with human data. On the other hand, there could also be artifacts that we did not rule out. In the worst case, the LLM could be attaining a high $R^2$ by learning heuristics specific to each \textit{rule}: although the specific object lists with which we compute $R^2$ were held out, all \textit{rules} were seen in tuning. In Experiment 4, we test the generalizability of tuning to unseen rules.

\section{Experiment 4: Generalizability of Tuning}
This experiment aims to characterize the generalizability of the tuning method used in Experiment 3, to investigate if rule-specific statistical heuristics are what drives the tuned LLM's high $R^2$ with human responses. To this end, we introduce withheld rules (not just object lists) from the language model's tuning process.

\subsection{Method}
\paragraph{Finetuning \& Data}

We use the same finetuning setup as Experiment 3. To rule out the possibility that the LLM is only learning some distributional statistic specific to each rule, we create another tuned model that is tuned only on 92 of 112 rules (i.e. excluding the training lists of 20 randomly chosen rules). We then evaluate the model on the object lists from these 20 held-out rules.

We also evaluate the model on rules that utilize an entirely different set of primitives than included in the base set of 112 rules. We take two additional rules proposed by \citet{ellisHumanlikeFewShotLearning2023} that involve minority/majority judgments; namely, the rules ``shape has the minority color" and ``shape has the majority color". \citet{ellisHumanlikeFewShotLearning2023} collected human learning data for these rules on one object list each.

\subsection {Results}
\begin{figure}[h!]
    \centering
   \includegraphics[width=0.6\textwidth]{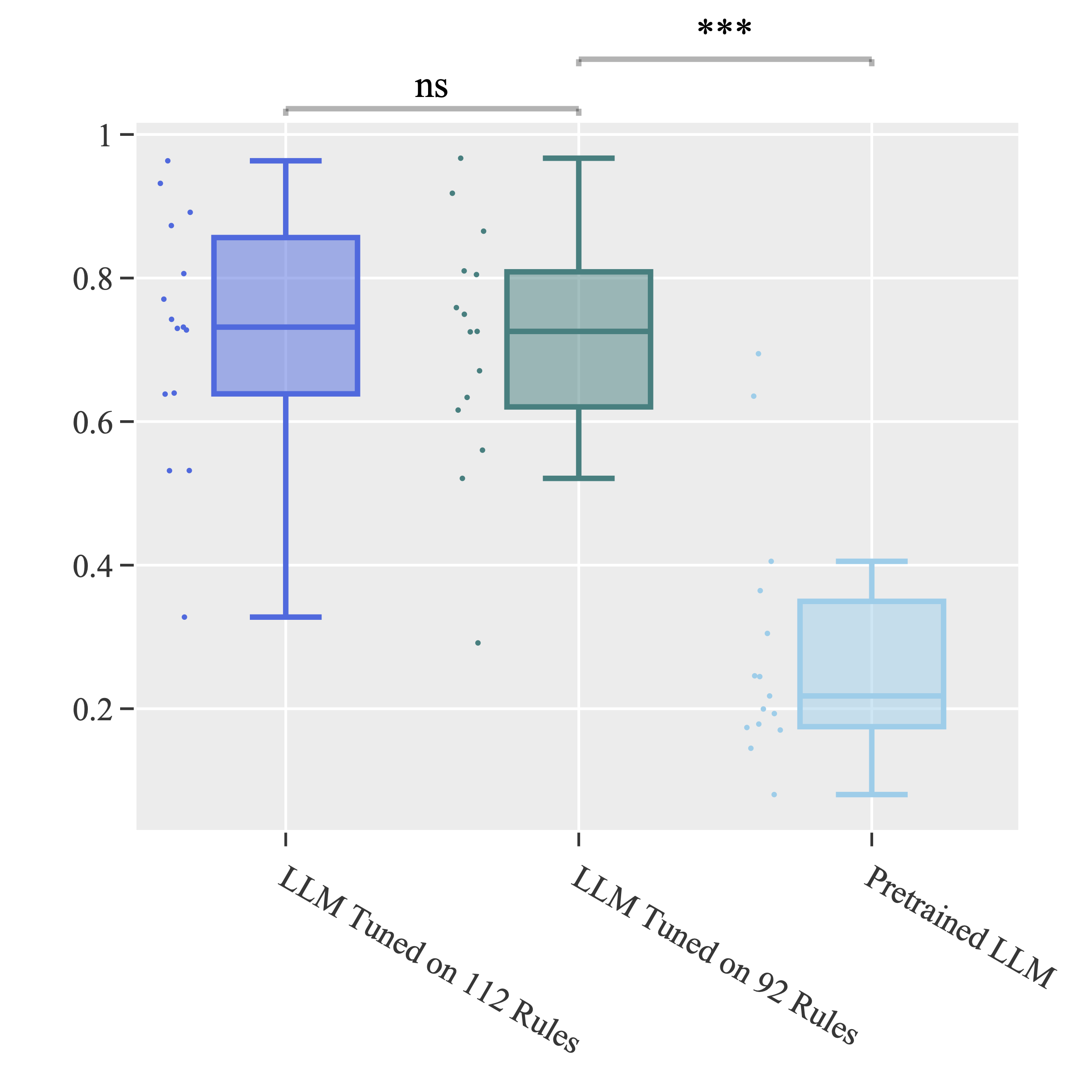}
  \caption{$R^2$ values between model and human posterior probabilities for responding \textit{True} between different variants of the LLM. Each scatter is one of the 20 randomly selected rules that were held out from one of the tuned LLM variants. Significance testing was done using two-tailed paired t-tests.} \label{fig:e4r2}
\end{figure}

\begin{figure}[h!]
    \centering
    \includegraphics[width=0.95\textwidth]{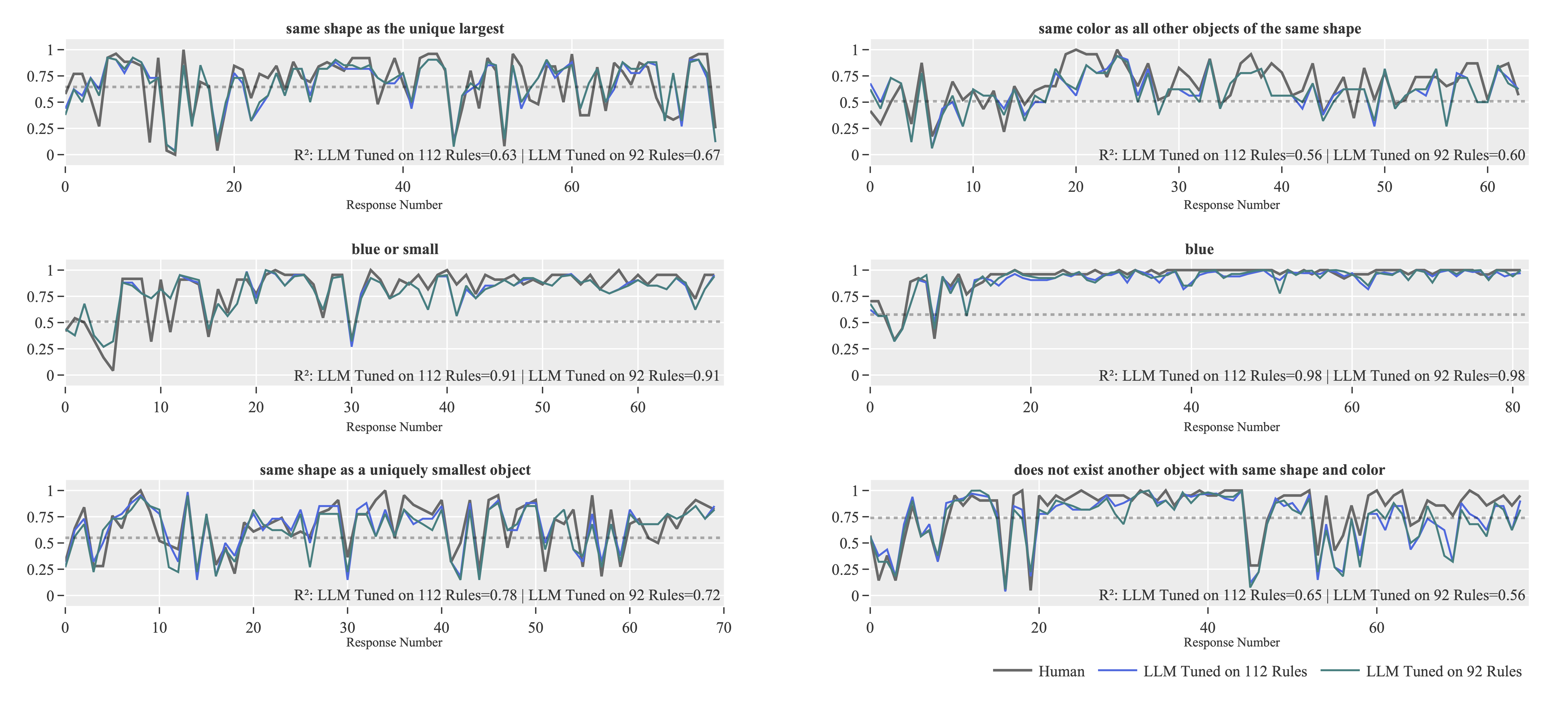}
    \caption{Accuracy trajectories on the held-out list. The figure shows the top, median and bottom two from the 20 held-out rules, ranked by difference between $R^2$ achieved by the LLM tuned on 112 rules and the LLM tuned on 92 rules. Dotted lines are the chance baseline for the rule, calculated as the accuracy from guessing `True' at the rate of the proportion of `True's in the object list.} \label{fig:trajectories-a}
\end{figure}

Figure \ref{fig:e4r2} compares the different LLM variants' $R^2$ with human responses on the 20 rules that were held out from one of the tuned LLM variants. We find that the $R^2$ values of the LLM that held out these 20 rules in tuning are, in fact, not significantly different than the LLM that had tuned on all rules. Furthermore, the LLM tuned on 92 rules still achieves significantly higher $R^2$ values than the pretrained LLM ($p < 0.0001$), although neither of these LLM variants had been tuned on these 20 rules. Qualitatively comparing the predictions of both tuned model variants specifically on these 20 held-out rules (Figure \ref{fig:trajectories-a}) reinforces that model trajectories are highly similar between the LLM tuned on all rules and the LLM tuned without the held-out rules. These results suggest that even without tuning with these 20 rules, the LLM is able to perform some transfer from the other 92 rules to maintain the same $R^2$ performance. This may be enabled by the fact that the 20 held-out rules share components with seen rules; for instance, tuning on ``\texttt{circle xor blue}" and ``\texttt{not (circle xor blue)}" may have enabled generalization to the unseen rule ``\texttt{circle xor (not blue)}".  A result that LLMs can generalize to new combinations of these constituents would itself be exciting, given prior work showing that LLMs have demonstrated poor compositional generalization capacity \citep{kimCOGSCompositionalGeneralization2020,hupkes2020compositionality}.

\begin{figure}[h!]
\includegraphics[width=\textwidth]{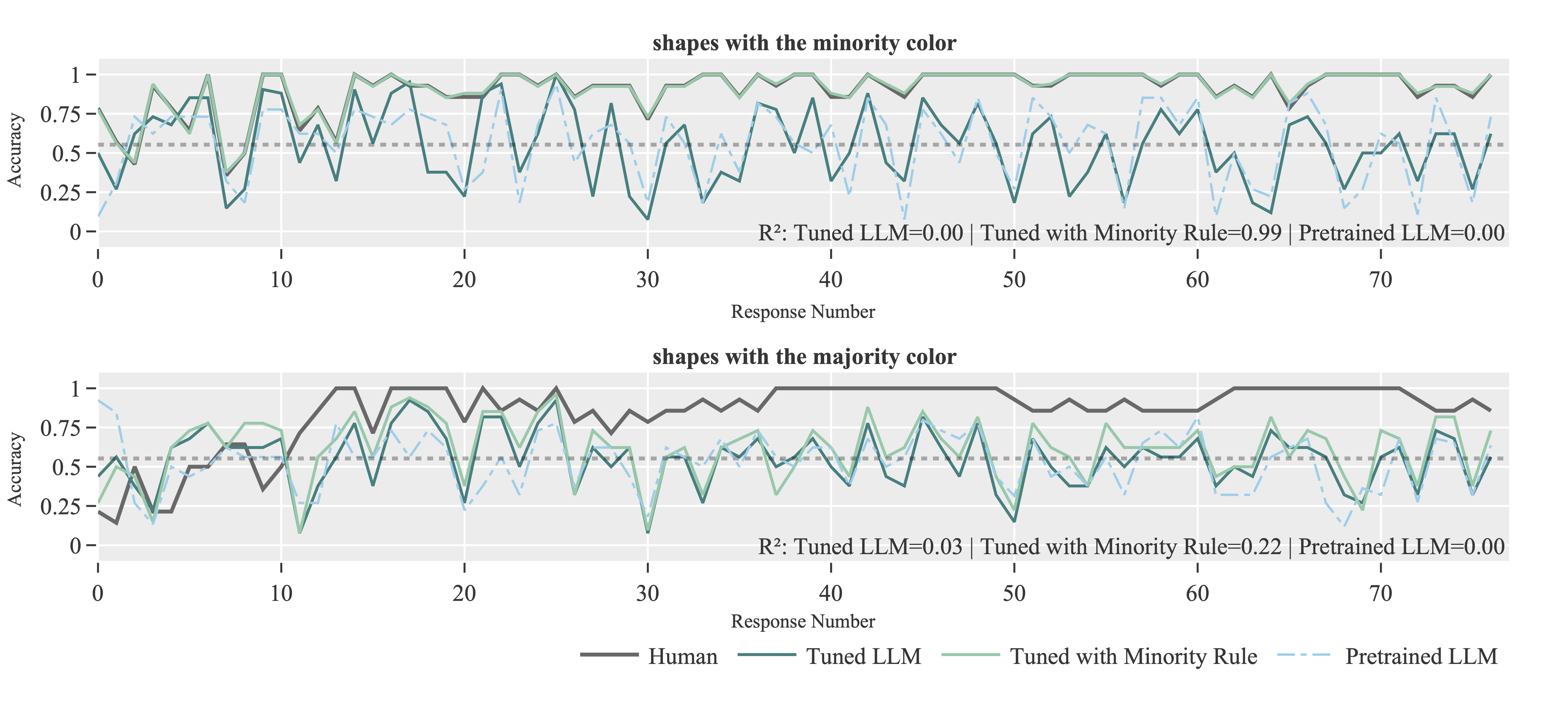}
    \caption{Accuracy trajectories for two additional rules and human data from Ellis (2023) that use majority/minority judgments, which is not used in any of the 112 rules from PTG16. Dotted lines are the chance baseline for the rule, calculated as the accuracy from guessing `True' at the rate of the proportion of `True's in the object list. The LLM tuned with the minority rule (light green) improves in $R^2$ on the majority color rule only very slightly over the LLM tuned only on PTG16's rules.}  \label{fig:trajectories-b}
\end{figure}

To isolate the effect of shared components, we now compare the tuned model's trajectory on the two additional rules from \citet{ellisHumanlikeFewShotLearning2023} that utilize majority/minority judgments,  which are not otherwise used in any of the prior 112 rules (Figure \ref{fig:trajectories-b}). In this setting, the tuned model behaves nearly identically to the pretrained model. Tuning the model on the rule involving minority concepts yields some, but minimal transfer to the model's performance on the majority concept, with $R^2$ improving from 0 to 0.22. The minimal amount of transfer could be because the LLM finds a weak relation between majority/minority concepts, because insufficient rules in the neighborhood of `majority' have been seen in tuning, or a combination of both.

\subsection{Discussion}
These results strongly suggest that tuning generalizes to both unseen object lists and unseen rules—as long as the unseen rules share the same abstract logical components. The sensitivity of tuning to logical components that are shared between rules, but not to the specific rules themselves, is hard to explain if the effect of tuning were only to teach the model surface heuristics. 

Instead, tuning could be teaching the LLM new concepts, or highlighting the relevance of existing concepts not previously brought to bear on the task. Tuning could cause latent concepts, already acquired in pretraining, to be re-weighted, effectively making the LLM's priors about the likelihood of certain concepts needing to be deployed in the task more human-like. While both possibilities are open, two considerations suggest the latter. First, the LLM does exhibit strong performance on the task even in its pretrained state, suggesting that tuning is more important for matching human learning trajectories than for discovering the right rules. Second, the tuned LLM remains around the chance baseline on some rules that humans do learn, which suggests that the required operators for those rules have not been acquired with tuning. We think that investigating what precisely changes in the model with tuning is a rich topic for future work.

\section{General Discussion}

As we argued in the introduction, LLMs' capacity for logical reasoning is relevant for cognitive theories of human reasoning if two conditions hold: (1) there is evidence that LLMs can model human behavioral data as well or better than existing (symbolic) accounts of logical reasoning from computational cognitive science and (2) there is evidence that the LLM represents a theoretically interesting alternative rather than “merely implementing” existing symbolic accounts. We consider each condition in turn.

\subsection{Do LLMs model Human Behavioral Data?}

Our results suggest that LLMs fit human reasoning behavior at least as well as existing theoretical accounts specified in terms of probabilistic inference over formal, symbolic logical operators. In four experiments, we tested whether LLMs can employ the same logical concepts humans do when inducing a rule from examples. We find that LLMs can perform the logical rule learning task in a way that closely matches human learners. Experiment 1 found that some LLMs achieve human-level accuracy on the logical rule learning task, across many rules of varying complexity. Using GPT4, an LLM that succeeded on the task at a rate comparable to humans, Experiment 2 found that it appeared to do so by applying hypotheses composed of logical concepts which it could verbalize and which tended to match the correct rule, a truth-functional equivalent, or a near neighbor. In Experiments 3 and 4, we gave a different LLM the same opportunity to tune free parameters to human responses as had been previously granted to Bayesian models. Experiment 3 found that a tuned LLM's learning trajectory matches humans better than any existing Bayesian model, with the LLM committing similar errors to human learners when given the same sequence of evidence. Experiment 4 showed that this result holds even when the LLM is tuned not only on different sequences of evidence, but also on entirely different rules as the ones they are tested on---but only as long as the two sets of rules share the same logical elements. 

The positive result that the LLM \textit{can} fit human learning trajectories runs counter to previous findings from \citet{ellisHumanlikeFewShotLearning2023} that LLMs on their own cannot fit human learners as well as when they are augmented with a Bayesian model. The biggest difference might be in our tuning of an open-weights LLM on human responses, instead of using a pre-trained LLM directly, as Ellis did. Since the time of writing, newer and stronger open-source models like Gemma 2 or Llama 3 have been released. We leave it to future work to test if these newer models could, possibly without tuning, achieve similar levels of correspondence with human learning signatures. 

The human-level accuracy found in Experiment 1 might have been achieved by LLMs finding some ineffable approximation to the correct rules, consistent with the ability of neural network to act as universal approximators \citep{hornik1989multilayer}. However, converging evidence across the next two experiments makes it more likely that these LLMs induced solutions that match the ones induced by human learners (cf. \citet{mccoyRightWrongReasons2019,mitchellAbstractionAnalogyMakingArtificial2021}). The high likelihood of the elicited hypotheses in Experiment 2 indicate that the LLM can combine logical concepts and feature primitives in meaningful ways to explain most of the provided evidence, while the high consistency of these self-reported hypotheses with the LLM's classification behavior suggests that it actually applied these logically-structured rules to label new objects. 

Since it is these hypotheses that yield similar accuracy scores to humans, especially for propositional rules, they could be plausible accounts of the rules formulated by human learners during the task. In Experiments 3 and 4, when given the same tuning to human responses as granted to Bayesian models, we found that an LLM's learning trajectory showed similar error patterns and learning curves to human learners at a very fine grain. In this, the LLM surpasses the best extant Bayesian models \citep{piantadosiLogicalPrimitivesThought2016, ellisHumanlikeFewShotLearning2023} in its ability to reproduce human logical rule learning. In nearly all cases where the LLM succeeded in learning the rule, we also find that it closely matched humans' learning trajectories. In contrast, a Bayesian model succeeded more often overall, but more often learned the rule by following a non-human-like learning trajectory (including some rules that humans generally failed to learn). When the LLM is consistently making the same mistakes in categorizing the same objects as human learners given the same stream of evidence, it is reasonable to infer that it is filtering through a similar set of candidate hypotheses with a similar search procedure as humans. Some existing literature supports the plausibility of this account: \citet{qiuPhenomenalPuzzlingTesting2024a} found LLMs are excellent hypothesis proposers, while \citet{xie2022an} have described a theoretical framework for how LLMs may be implicitly implementing Bayesian inference over learned concepts when performing in-context learning. Note that the close match between LLMs' and humans' learning trajectories for particular rules also implies that the LLMs are testing hypotheses articulated over already-known primitives (as humans must be doing), and not inducing the primitives and the rules articulated over these primitives simultaneously.

LLMs—like humans—did not learn every kind of rule equally well. The easiest rules for LLMs to learn were those formulated in propositional logic, using only the Boolean operators. These were also the rules that were the easiest for humans to learn, and the rules on which the learning trajectories taken by LLMs and humans matched most closely. On the other hand, LLMs struggled to learn rules articulated over the existential and universal quantifiers of first-order logic. Human accuracy on those rules was much lower as well, and the correspondence of learning trajectories between humans and LLMs was weaker. Of course, every successfully learned rule is (at least truth-conditionally) alike, while every failed attempt can fail in its own way. It is therefore possible that the differences between humans and LLMs on FOL rule-learning mask an underlying similarity: LLMs avoided forming rules with quantifiers, preferring to concatenate their earlier hypotheses with additional Boolean compositions of basic features. Whether humans take the same strategy is unclear, and attempts to get human participants to articulate what rule they are using have previously failed \citep{piantadosiLogicalPrimitivesThought2016}.

However, PTG16 did find that although a Bayesian model with propositional logic could fit human responses on propositional rules extremely well, a Bayesian model equipped with the existential and universal primitives of first-order logic was unable to provide as good a fit to human responses on rules that required first-order logic. In contrast, we found that an LLM fit human responses on these rules better, even as it did worse in actually learning FOL rules. If human participants were employing a strategy similar to that of LLMs—of forming alternative rules using extended Boolean compositions—it would account for the weaker fit of the Bayesian model equipped with first-order logic to human responses on FOL rules, and for the relatively better fit of LLMs to humans on these rules. In that case, it is possible that the logical rule learning task, at least in the form in which it has been used, can fail to induce participants, either human or LLM, to formulate rules that involve quantifiers. If that is the case, this task is unsuited for testing the representation and deployment of these particular logical operators, and the possibility that LLMs possess the corresponding logical concepts remains open. The same might hold for more complex logical operators, such as the second-order quantifiers that correspond to the meanings of logical words like \textit{most}.

\subsection{Do LLMs ``Merely Implement'' Existing Symbolic Models?}

Given results that show LLMs to be the most empirically successful model of human behavior on this task, the second important question for theories of human cognition is whether LLMs ``merely implement'' existing symbolic accounts of formal logical concepts, or whether they might offer new theoretical insights? On this question, our experiments provide highly suggestive evidence, and further work will be required before we can fully understand the implications of LLMs for theories of the logical (or logic-like) operators that underlie human reasoning. 

In considering this question in the context of this task, it is necessary to distinguish between the primitive logical concepts themselves and the inference procedure which operates over those concepts to select a hypothesis that is composed of these primitives, given the evidence seen up to that point. Our focus in this paper has been primarily on the former: the logical primitives of human thought. The Bayesian pLoT model, as an architecture, is primarily a description of the latter. It is an inference procedure which can be deployed over any library of primitive concepts so long as they are precisely defined \textit{a priori}. As of PTG16, the best model of human behavior was this inference procedure applied to the logical primitives of boolean and first-order logic, and this is the model that has been our focus and primary contrast with the LLM.

In comparing the LLM to the Bayesian pLoT model, therefore, primitive concepts and inference procedure are entangled, and there are multiple interpretations that are consistent with the experimental results we have presented here. One possibility is that the LLM and the pLoT model (and by extension, humans) all employ the same primitive logical operators, but that the LLM and pLoT model differ in the inference procedures they use to compose these primitives into rules in order to perform the task. Another possibility is that both the LLM and the pLoT model employ the same inference procedure, but do so over different sets of primitive operators. A third option is that the LLM and pLoT differ along both dimensions.

The experimental evidence we present here is inconclusive with respect to these alternatives. Our evidence from the rule elicitation in Experiment 2 suggests that, when verbalized using standard logical operators, there is a qualitative difference in the rules that are inferred. However, given that the LLM had to verbalize its rule in English, we cannot tell from the surface form of the rule alone whether the operators it describes are the standard logical ones or not -- does the \textit{or} in ``\textit{large or blue}'' mean boolean OR or something somewhat close to OR, but a with different inferential role or domain of application? Similarly, the higher correlations of the LLM than the pLoT with human trajectories in Experiments 3 and 4 are inconclusive on this question. The higher correlation may be due to the LLM having more human-like hypotheses (that is, composed of more human like operators) or to the LLM employing more human-like hypothesis-updating in light of new (inconsistent) evidence, or both. Without further studies, we cannot say which of these is the most likely.

While the evidence is inconclusive on the issue of \textit{what} is more human-like—the primitives, the inference procedure, or both—the evidence does strongly indicate that \textit{something} is different between the LLM and the pLoT, and that something about the LLM models humans' thinking in a way that the pLoT does not. That is, it would be unlikely to observe both improved fits and qualitative differences in the verbalized rules if the LLM represented a ``mere implementation'' of the Bayesian pLoT itself. It seems likely, then, that directly investigating (and distinguishing) both the primitives and the inference procedure that LLMs use in tasks such as this will prove relevant to cognitive theories of human reasoning. 

The best evidence for what kind of logical (or logic-like) operators LLMs have may come from mechanistic interpretability studies of the representations that LLMs use in this task \citep{mcgrath2024DNNsInformTheory}. Identifying the circuits that LLMs use to subsequently verbally express rules in terms of operators like \textit{and, or} and \textit{not} can provide further hypotheses about how neural circuits might implement the corresponding concepts in human productive, combinatorial thought, and how this might correspond to, or differ from, the relatively simple wiring of logic gates in classical computing.

\section{Conclusion}

Across four experiments, we found that LLMs can learn and deploy human-like logical concepts, use them to compose and verbalize logically structured thoughts, and apply and revise these as running hypotheses about latent rules that fit observed evidence. Such logical concepts and thoughts have long been a flagship example of the type of higher cognition that requires a compositional Language of Thought, which has been argued to be a fundamentally different architecture from that of artificial neural networks that are not specifically designed to implement logic or compositionality \citep{fodorConnectionismCognitiveArchitecture1988a,quilty-dunnBestGameTown2023a}.

In contrast to this, our findings argue that neural networks deserve consideration as empirically adequate models of logical thought. And if having logical concepts is one of the \textit{constitutive} properties of a language of thought \citep{mcgrath2023properties}, then our findings also suggest that certain large language models meet this criterion.

At the same time, our findings do more than suggest that LLMs can fit pre-existing theories of cognition. Given that the large language models we tested actually fit how humans learned logically structured rules better than symbolic models endowed with logical primitives, these models may not just implement existing theories of logical thought. Instead, they may open up new opportunities to discover differences between formal and human logic, serving as a computational testbed for the study of a key aspect of human cognition.

\section{Contribution}
\textbf{Alyssa Loo}: Conceptualization, Methodology, Formal analysis, Data curation, Software, Visualization, Writing - original draft; \textbf{Ellie Pavlick}: Conceptualization, Supervision, Writing - original draft, Writing – review and editing; \textbf{Roman Feiman}: Conceptualization, Supervision, Writing - original draft, Writing – review and editing

\section{Acknowledgements}
We thank Jake Russin, Aaron Traylor, Michael Lepori and Apoorv Khandelwal for insightful comments that greatly enriched this project. We also thank Steven Piantadosi for providing direction on using the Fleet library used in PTG16, and Kevin Ellis for the helpful discussion about his prior work with LLMs on this task.  

\section{Funding Sources}
This research did not receive any specific grant from funding agencies in the public, commercial, or not-for-profit sectors.

\bibliographystyle{elsarticle-harv} 
\bibliography{cas-refs-2}

\clearpage
\appendix

\section{Prompting Method}
\label{sec:promptmethod}
\subsection{API Details for Model Data Collection}
We collected data for GPT4 and GPT3.5 by querying the \texttt{gpt4-1106-preview} and \texttt{gpt-3.5-turbo-1106} at $T=0.7$ in April 2024. We collected data for Mixtral 8x7B Instruct and the chat and completion models of Llama2 (70B) by querying Together AI's API endpoints also at $T=0.7$.
\clearpage
\subsection{Prompt Templates}
\begin{lstlisting}[caption={Formatted for a \textbf{completion} model for Experiment 2, where the model has been shown the labels for Group 1 and is being queried for its label on the first object of Group 2. performing inference on Group 2. Completions are extracted object-by-object with labels revealed on previous objects.}, captionpos=b]
# Instructions
Learn the secret rule to label the objects in groups correctly. The rule may consider the color, size and shape of objects, and may also consider the other objects in each group. If an object in a group follows the rule, it should be labeled 'True'. Otherwise it should be labeled "False".

# Quiz
## Group 1
small blue rectangle
medium green triangle
large blue circle

## Group 1 Answer Key
small blue rectangle -> True
medium green triangle -> True
large blue circle -> False

## Group 2
large green circle
small yellow triangle

## Group 2 Answer Key
large green circle ->
\end{lstlisting}

\clearpage
\begin{lstlisting}[caption={Formatted for a \textbf{chat} model for Experiment 1, where the model has been shown the labels for Group 1 and is being queried for its labels on Group 2. Labels are extracted set-by-set with labels revealed on previous sets.}, captionpos=b]
{`role': `system': `content': "Learn the secret rule to label the objects in groups correctly. The rule may consider the color, size and shape of objects, and may also consider the other objects in each group. If an object in a group follows the rule, it should be labeled `True'. Otherwise it should be labeled `False'. Be concise and always follow the arrow `->' with a object's label."}

{`role': `user', `content': "Label the following objects in Group 1:
- small blue rectangle
- medium green triangle
- large blue circle"} 

{`role': `assistant',
'content': f``The labels for the objects in Group 1 are:
- small blue rectangle -> True
- medium green triangle -> False
- large blue circle -> True"}

{`role': `user', `content': "Label the following objects in Group 1:
- large green circle
- small yellow triangle}
\end{lstlisting}

\clearpage
\begin{lstlisting}[caption={Prompt format for Experiment 2, where the model has been shown the labels for Group 1 and is being queried for a rule elicitation and its labels on Group 2. Labels are extracted set-by-set with labels revealed on previous sets.}, captionpos=b]
{`role': `system': `content': Learn the secret rule to label the objects
in groups correctly. The rule may consider the color, size and shape
of objects, and may also consider the other objects in each group. 
If an object in a group follows the rule, it should be labeled `True'. 
Otherwise it should be labeled `False'. Begin your answer with 
'Based on your examples, I've learned that...' and give your 
best concise description of the rule without asking for more evidence. 
Make sure your rule is consistent with the labels for all objects 
so far. Be concise and always follow the arrow `->' with an object's 
label. Label ALL objects in the given group.}

{`role': `user', `content': "The labels for the objects in Group 1 are:
- small blue rectangle -> True
- medium green triangle -> True
- large blue circle -> False

Now concisely describe the labeling rule based on these groups of objects, 
then label all the following objects in Group 2:
- large green circle
- small yellow triangle"}
\end{lstlisting}

\subsection{Comparison with Ellis}
\label{sec:elliscomparison}
Our results observe better success in rule-learning with GPT-4 than previously reported by \citet{ellisHumanlikeFewShotLearning2023}.The prompt used in our study vary from Ellis' prompt design in a few ways: labels of previous sets are formatted as `assistant' turns, rather than as part of the task description in the user message; labels are elicited as part of an bulleted list instead of as an answer to a natural language question; objects are listed in bullet points and followed by their labels (“large yellow triangle -> True”) instead of grouped according to their labels  (“POSITIVE EXAMPLES: (large yellow triangle) (medium blue circle)”); and lastly, our prompt preamble does not give examples of possible concepts while Ellis’ does. Furthermore, although both studies query the endpoint designated as `GPT-4',  the behavior of closed-source models has been to shown to be able to change drastically even in a short period of time \citep{Chen2024How}. Model updates between the time of Ellis' study and our queries submitted in April 2024 may have contributed to the difference in observed performance.

\section{$R^2$ with GPT-4}
\label{sec:gpt4r2}

Considering that GPT-4 was a top-performing model in Experiment 1 and the model of focus in Experiment 2, it is reasonable to be curious about its $R^2$ with human responses as measured for other models in Experiment 3. We thank our reviewer for pointing out that this would be an interesting datapoint.

At the time when we collected data for these experiments in April 2024, we did not query for log-probabilities from the GPT-4 endpoint that are required to compute the $R^2$ measure with human responses. We attempted to reproduce our results with the same \texttt{gpt-4-1106-preview} endpoint in March 2025, now storing the log-probabilities, but we could not reproduce its previous rule learning performance. This is likely because closed-source changes were made to the model in the interim. This means that it is not possible to get $R^2$ measures on the same GPT-4 results from April 2024 that we report in Experiment 1. 

For completeness, we report here the results from the most recent endpoint with $R^2$ calculated from its output log-probabilities. 

\paragraph{Method} To calculate the log-probabilities, we sum the probability mass of all `True` and `False` tokens (up to variance in whitespace and capitalization) that appear in the top ten token choices at the labeling position, with sampling temperature 0.7. We normalize these raw summed masses for `True' and `False' against their total mass to form a probability distribution.

The system instruction is slightly modified from that of the original April 2024 query, as the latest model had a higher refusal rate when information was insufficient. We include a line encouraging an answer even if there was partial information.

\begin{lstlisting}[caption={System instruction for GPT-4 queries in March 2025}, captionpos=b]
{`role': `system': `content': "Learn the secret rule to label the objects in groups correctly. The rule may consider the color, size and shape of objects, and may also consider the other objects in each group. If an object in a group follows the rule, it should be labeled `True'.  Otherwise it should be labeled 'False'. Your response should concisely describe your best guess of the rule, then restate each object with an arrow '->' and then followed by a True or False label. Always follow the arrow '->' ONLY with each object's label, and nothing else. The label should be always either be 'True' or 'False'. You will receive more information after every set you label, so always just make your best guess even if there is only partial evidence. Good luck!"}
\end{lstlisting}

\paragraph{Results} The task performance of GPT-4 queried more recently in March 2025 is weaker than those from April 2024, especially for FOL rules. The $R^2$ is about comparable to that of the pretrained Gemma-7b checkpoint at 0.484. 

\begin{center}
\begin{table}[!h]
  \scriptsize
  \centering
  \begin{tabular}{p{9em}|C{3.8em}|C{3.8em}|C{3.8em}|C{3.8em}|C{3.8em}|C{3.8em}}
  \bottomrule
  & \multicolumn{2}{c|}{\textbf{All Rules}} & \multicolumn{2}{c|}{\textbf{Propositional Rules}} & \multicolumn{2}{c}{\textbf{FOL Rules}} \\
  \toprule 
  Model & Overall & Last Quarter & Overall & Last Quarter & Overall & Last Quarter \\
  \midrule
  \textit{Chat Models}  & & & & &\\
   GPT4 & \underline{0.764} & 0.813 & \textbf{\underline{0.908}} & \textbf{\underline{0.987}} & 0.701 & 0.737 \\
 GPT4 (March 2025) & 0.730 & 0.780 & \underline{0.889} & \textbf{\underline{0.987}} & 0.662 & 0.693 \\
Gemma-7b & \textbf{\underline{0.795}} & \textbf{\underline{0.840}} & \underline{0.886} & \underline{0.969} & \textbf{\underline{0.756}} & \textbf{\underline{0.784}} \\
GPT2-XL* & 0.654 & 0.684 & 0.663 & 0.722 & 0.650 & 0.667 \\
GPT2* & 0.622 & 0.707 & 0.634 & 0.703 & 0.617 & 0.709 \\
\midrule
Human (25 Sets) & $0.774 \pm 0.0109$ & $0.835 \pm 0.0148$ & $0.858 \pm 0.106$ & $0.932 \pm 0.0235$ & $0.737 \pm 0.0132$ & $0.792 \pm 0.0187$\\
Human* (14 Sets) & $0.726 \pm 0.0110$ & $0.798 \pm 0.0170$ & $0.798 \pm 0.020$ & $0.900 \pm 0.028$ & $0.695 \pm 0.0132$ & $0.753 \pm 0.021$\\
  \bottomrule
  \end{tabular}
  \caption{Performance of GPT4 queried in March 2025 compared to other models initially reported in Experiment 1. Model averages that are above the lower SD bound of human averages for the corresponding type of rule are underlined. The highest scores for each type of rule are bolded.}
\label{tab:rq1results-mar2025}
\end{table}
\end{center}

\begin{figure}[h!]
\centering
\includegraphics[width=\textwidth]{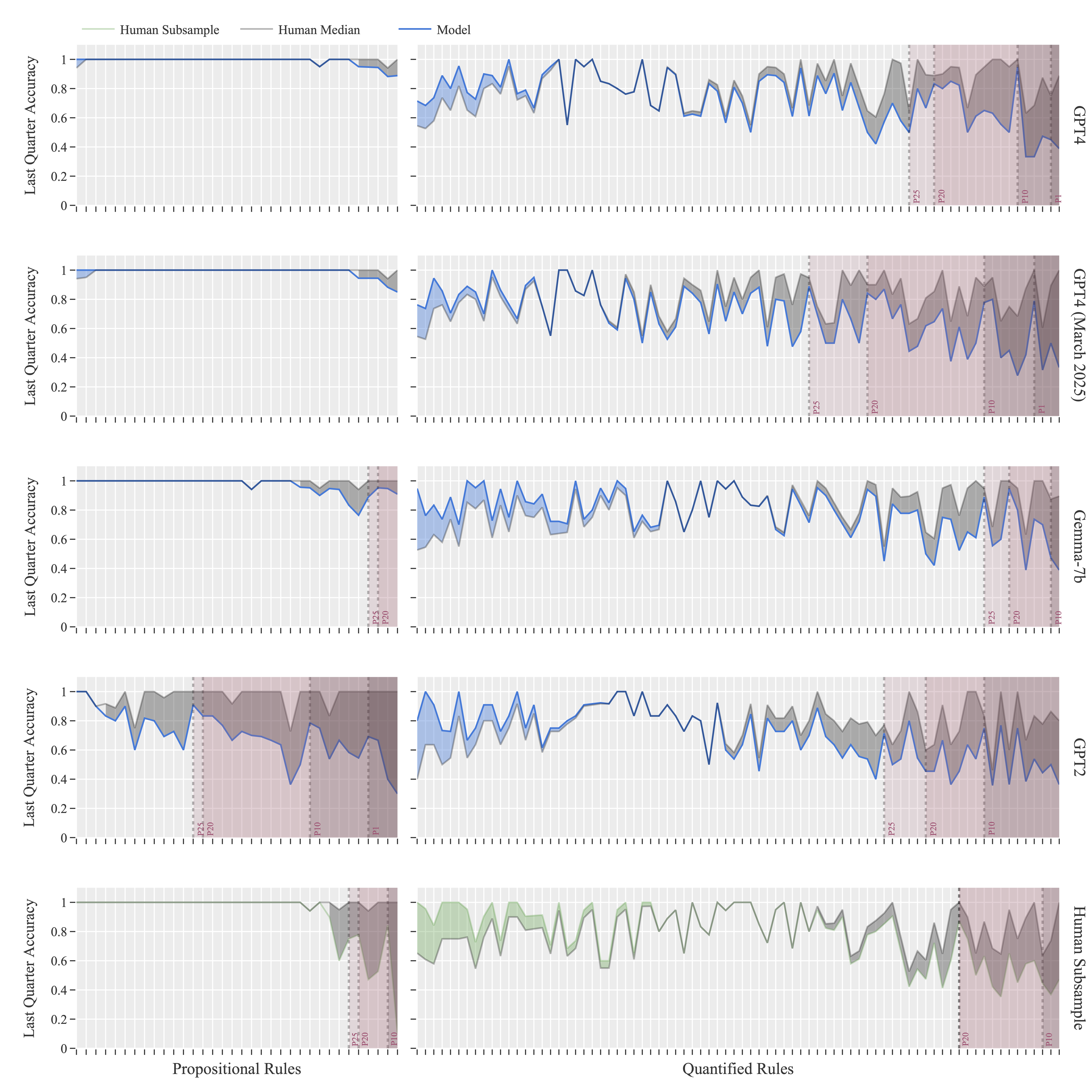}
\caption{Difference between accuracy scores between each model and the human median, with GPT-4 (March 2025) added. Each $x$-value is a rule with two $y$-values plotted: the median last-quarter accuracy for human particpants (gray) and for the model in the given row (blue). Shaded red regions show the intervals of rules where model accuracy falls below the 25$^{th}$, 20$^{th}$, 10$^{th}$ and 1$^{st}$ percentile of human last-quarter accuracies.}
\label{fig:deltagraph_with_gpt4new}
\end{figure}

\begin{figure}[h!]
    \centering
    \includegraphics[width=0.5\linewidth]{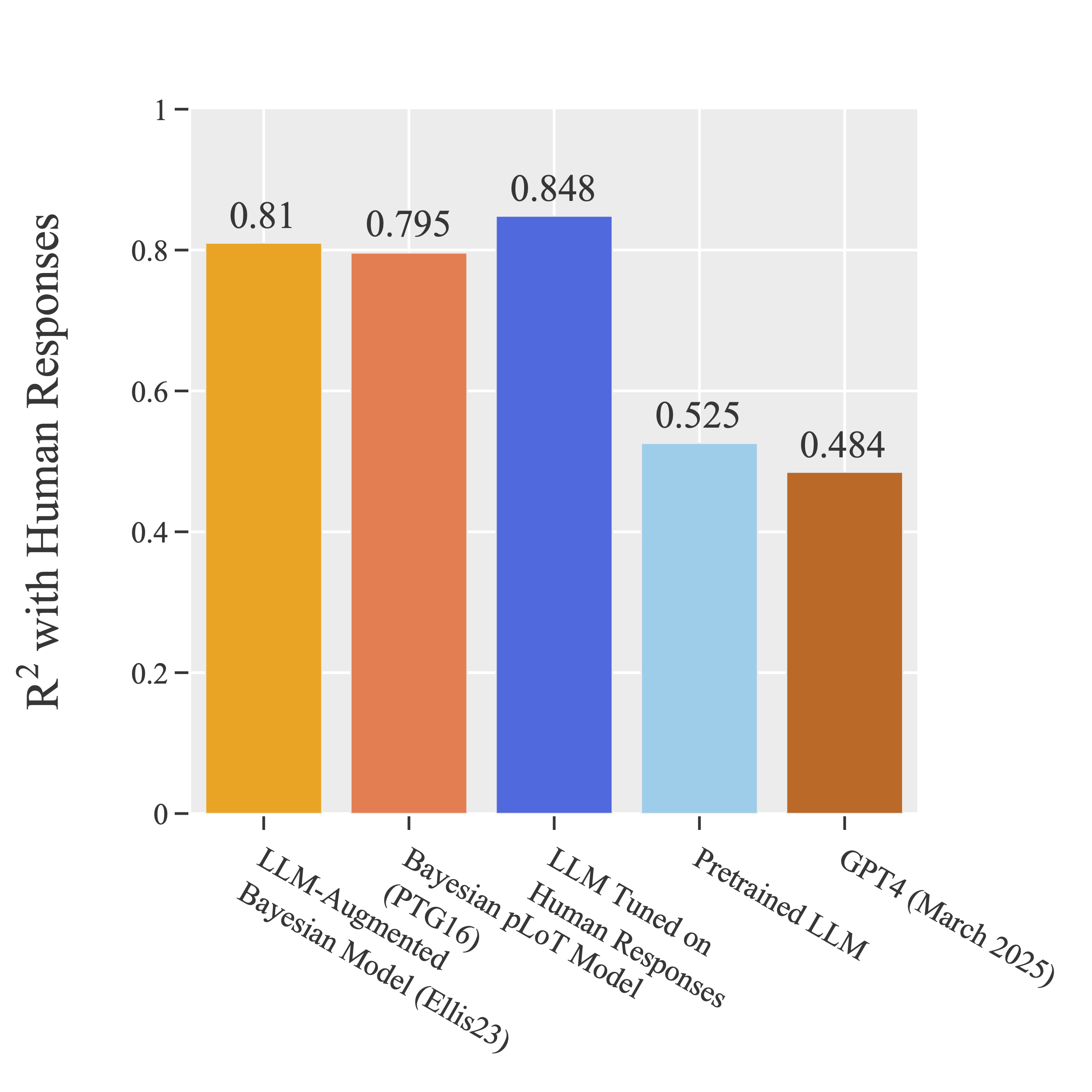}
    \caption{$R^2$ values between model and human posterior probabilities for responding \textit{True} between different models, with GPT-4 (March 2025)} \label{fig:r2a_gpt4new}
\end{figure}
  
\clearpage
\section{Finetuning Details}
\label{sec:ftmethod}
Finetuning was done on two Quadro RTX 6000 GPUs. We loaded pretrained Gemma (7B) at 4-bit precision using NF4 quantization with double quantization, with a 16-bit compute datatype using \texttt{bitsandbytes}, a wrapper library that implements quantization methods from \citet{dettmers2022llm, dettmers2023qlora, dettmers2023spqr, dettmers2023case, dettmers20218bit}. We apply rank-stabilized Q-LoRA on all linear modules with $r=64$, $\alpha=32$, dropout at $0.05$, and without updating biases, using the \texttt{PeftModel} class \citep{peft}. Training was done with the Hugging Face \texttt{Trainer} class, using paged 8-bit Adam as the optimizer with 30 warmup steps, a learning rate of $1e^{-5}$, batch size of 1, gradient accumulation over every 4 steps and gradient checkpointing. 

The LLM tuned on 112 rules to human responses as loss was fine-tuned for 72 epochs taking 4.5 hours, with training loss decreasing from 0.1967 to 0.0008. The LLM tuned to the correct answers with cross-entropy loss was fine-tuned for 72 epochs taking 2.5 hours, with training loss decreasing from 11.752 to 11.494. The LLM tuned with 20 heldout rules was fine-tuined for 72 epochs taking 1 hours, with training loss decreasing from 0.1572 to 0.0007.

\begin{table}
\section{Additional Rule Elicitations}
\label{sec:fol-attempts}
\centering
\renewcommand{\arraystretch}{1.7}
\small
\begin{tabular}{l|p{11em}|c|p{11em}|c}
\# & \textbf{True Rule} & \textbf{Set \#} & \textbf{LLM Rule} & \textbf{$\mathcal{L}$}\\
\toprule
1 & one of the largest or smallest &  6 & another object in the group shares two features & 0.11\\
2 & one of the largest or smallest & 13 & at least one blue object and at least one green object in group & 0.62 \\
3 & exactly one blue object in the set & 10 & unique size in group & 0.56\\
4 & does not exist another object with same shape and color & 19 & shares an attribute with some other object in its group & 0.69 \\
5 & exactly one blue object that is the same shape & 13 & shape and color have been previously labeled true & 0.62
\end{tabular}
\caption{Examples of where the model used first-order logic in its elicited rules for some object sets from the annotated subset of FOL rules. Likelihood ($\mathcal{L}$) describes the proportion of known labels that the rule correctly accounts for.}
\end{table}

\begin{figure}
\section{Additional Learning Trajectories}
\label{sec:additional-traj}
\centering
\includegraphics[width=0.94\textwidth]{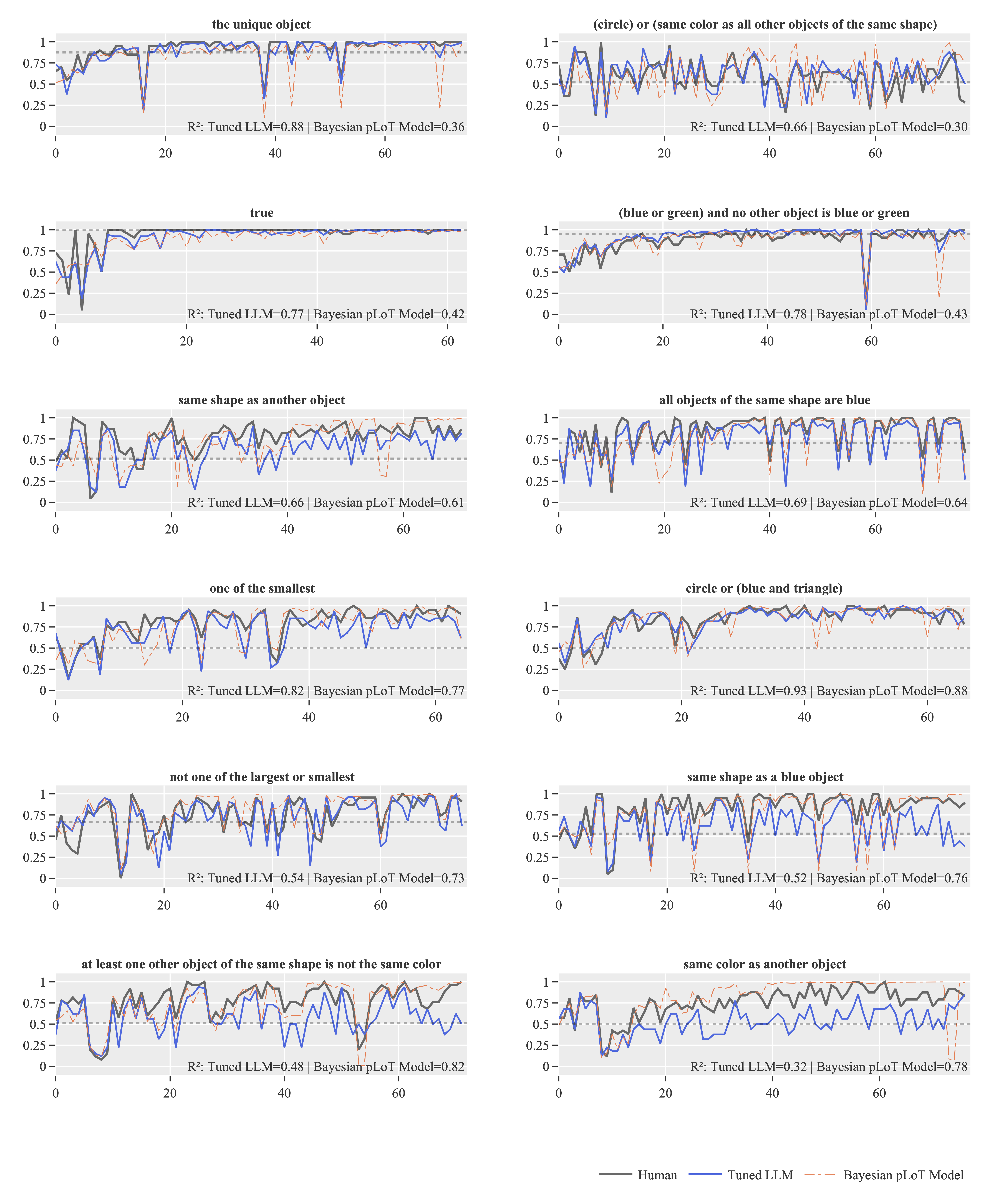}
\caption{Accuracy trajectories for the Tuned LLM and Bayesian pLoT models.  The $y$-axis plots accuracy and $x$-axis plots response number. The figure shows the top, median and bottom four rules, ranked by the difference between $R^2$ achieved by the Tuned LLM and the Bayesian pLoT model. Dotted lines are the rule's chance baseline, calculated as the accuracy from guessing `True' at the rate of the proportion of `True's in the object list.}
\end{figure}

\clearpage
\section*{Figure Captions}
\begin{enumerate}
\item Title: Logical rule learning task.\\Caption: Sample inputs compare the original presentation to human participants in Piantadosi, et al. (2016) (`PTG16') versus our adapted format for a chat completion model. Inputs are constructed by concatenating instruction text with an object list and its corresponding labels according to the rule. In this example, the rule is `\texttt{same shape as a yellow object}'
\item Title: Difference between accuracy scores between each model and the human median. \\ Caption: Each $x$-value is a rule with two $y$-values plotted: the median last-quarter accuracy for human participants (gray) and for the model in the given row (blue). The inter-graph fill represents the difference between the last-quarter accuracies over an interval of rules: blue indicates that the model's accuracy is higher than that of the human median in that interval and gray vice versa. Within each row, rules are sorted in descending order of the accuracy difference for that model. Shaded red regions show the intervals of rules where model accuracy falls below the 25$^{th}$, 20$^{th}$, 10$^{th}$ and 1$^{st}$ percentile of human last-quarter accuracies. An analogous human subsample (green) is constructed by taking a randomly selected human participant's score for each rule. Rules are split into rule type by column.

\item Title: $R^2$ values between model and human posterior probabilities for responding \textit{True} between different models.\\Caption: (a) $R^2$ values across all objects from all rules. (b) $R^2$ values plotted where each scatterpoint is a rule. Significance testing was done using two-tailed paired t-tests.
\item Title: Learning trajectories compared between human participants, the tuned LLM, and the pLoT model with a FOL grammar from PTG16.\\Caption: The two rules were chosen to be representative of a good and bad fit for the Tuned LLM compared to the pLoT model. The gray dashed line shows the chance baseline, defined as the accuracy from guessing `True' at the empirical rate of `True' labels for the exemplar list.

\item Title: How each model's $R^2$ with human responses changes with the model's last-quarter accuracy on each rule.\\Caption: Quadrant annotations are guides for qualitative interpretation. First, the LLM has overall higher $R^2$ with humans. Second, on rules where the tuned model achieves high accuracy, it tends to also have a high $R^2$ with human responses; when the Bayesian model achieves high accuracy, it may still have a low $R^2$ with humans.

\item Title: Comparison of the relationship between accuracy and $R^2$ between the Bayesian model and tuned LLM.\\Caption: The LLM tuned on answers attains a higher last-quarter accuracy on the task than that tuned on human responses, but it has a poor correlation with human responses. Its improvement in correlation over the pretrained model can be interpreted as the correlation accounted for by learning rules correctly.

\item Title: $R^2$ values between model and human posterior probabilities for responding \textit{True} between different variants of the LLM.\\Caption: Each scatter is one of the 20 randomly selected rules that were held out from one of the tuned LLM variants. Significance testing was done using two-tailed paired t-tests.

\item Title: Accuracy trajectories on the held-out list. The figure shows the top, median and bottom two from the 20 held-out rules, ranked by difference between $R^2$ achieved by the LLM tuned on 112 rules and the LLM tuned on 92 rules.\\Caption: Dotted lines are the chance baseline for the rule, calculated as the accuracy from guessing `True' at the rate of the proportion of `True's in the object list.

\item Title: Accuracy trajectories for two additional rules and human data from Ellis (2023) that use majority/minority judgments. Caption: Minority/majority concepts were not used in any of the 112 rules from PTG16. Dotted lines are the chance baseline for the rule, calculated as the accuracy from guessing `True' at the rate of the proportion of `True's in the object list. The LLM tuned with the minority rule (light green) improves in $R^2$ on the majority color rule only very slightly over the LLM tuned only on PTG16's rules.

\item Title: Accuracy trajectories for the Tuned LLM and Bayesian pLoT models. \\ Caption: The figure shows the top, median and bottom four rules, ranked by the difference between $R^2$ achieved by the Tuned LLM and the Bayesian pLoT model. Dotted lines are the chance baseline for the rule, calculated as the accuracy from guessing `True' at the rate of the proportion of `True's in the object list.
\end{enumerate}

\end{document}